\documentclass{article} % For LaTeX2e
\usepackage{iclr2027_conference,times}
\usepackage{amsmath,amsfonts,bm}

\def\eqref#1{equation~\ref{#1}}
\def\1{\bm{1}}

\DeclareMathAlphabet{\mathsfit}{\encodingdefault}{\sfdefault}{m}{sl}
\SetMathAlphabet{\mathsfit}{bold}{\encodingdefault}{\sfdefault}{bx}{n}

\usepackage{hyperref}
\usepackage{url}
\usepackage{graphicx}
\usepackage{booktabs}
\usepackage{subcaption}
\usepackage{multirow}
\usepackage{bm}
\usepackage{placeins}

\title{Self-Reconstruction Dynamics for Autoencoder Reconstruction Refinement}
\author{Hitoshi ~Iyatomi \\
Department of Applied Informatics, Faculty of Science and Engineering\\
Hosei University\\
Tokyo, 184-8584, Japan \\
\texttt{iyatomi@hosei.ac.jp} \\
}

\iclrfinalcopy % Uncomment for camera-ready version, but NOT for submission.

\begin{document}
\maketitle
\lhead{Under review as a conference paper at ICLR 2027}
\begin{abstract}
    Standard autoencoder (AE) inference uses a single encoder--decoder pass,
although the resulting latent representation need not be optimal for each
sample under the fixed decoder.
This raises a natural question:
\emph{Can a trained AE itself reveal information useful for improving its own reconstruction?}
We investigate this question by repeatedly applying a frozen AE to its own reconstruction, producing transient image- and latent-space trajectories that we call \emph{Self-Reconstruction Dynamics (SRD)}.
Although repeated self-reconstruction degrades fidelity in the AEs studied here, the resulting SRD contains useful sample-specific information for correcting the initial reconstruction.
We therefore propose \emph{SRD-guided Reconstruction Refinement (SRD-RR)}, which predicts a latent correction from a short SRD while keeping the AE frozen and requiring no per-sample test-time optimization.
We also introduce an \emph{MSE recovery ratio (MSE-recov)} relative to an empirical decoder-optimized reference.
Across six image datasets, SRD-RR recovers an average of 38.6\% of the empirically recoverable MSE gap with one trajectory transition and 45.3\% with two.
A two-transition variant trained without direct access to the original images, using an SRD-derived pseudo-target, achieves 40.7\% recovery and a 1.74\,dB average PSNR gain.
Removing trajectory information substantially reduces the gain, while cross-sample trajectory assignment causes severe degradation, showing that the useful SRD information is strongly sample-specific.
Nonlinear SRD-conditioned refinement also consistently outperforms both fixed and trained linear latent correction.
We further evaluate SRD on a pretrained DINOv2-based representation autoencoder (RAE) with substantially different latent dynamics.
SRD conditioning again improves a matched trajectory-free predictor, showing that trajectory information remains useful in this different AE setting.
However, pixel-MSE latent refinement exposes a strong mismatch between pixel fidelity and perceptual quality, while the SRD-derived pseudo-target substantially mitigates this perceptual degradation.
Together, these results establish SRD as a useful sample-specific signal for reconstruction refinement, while showing that the choice of refinement objective determines how this information translates into pixel and perceptual quality.
\end{abstract}
\section{Introduction}
    %\section{Introduction}
%\vspace{-5pt}
Autoencoders (AEs) map high-dimensional observations to latent representations
and reconstruct them through a decoder \citep{hinton2006reducing}.
Reconstruction quality remains important in modern latent generative models,
including VAEs and latent diffusion models
\citep{kingma2013auto,rombach2022high}, while recent representation autoencoders (RAEs) emphasize both reconstruction fidelity and representation quality \citep{zheng2026diffusion}.
Standard AE inference is essentially one-shot: an encoder predicts a latent representation that is immediately decoded.
Yet this representation need not yield the best reconstruction of each sample under the fixed decoder.
Latent inversion and learned inference refinement likewise show that sample-specific refinement can improve an initial encoder prediction
\citep{zhu2016generative,xia2022gan,cremer2018inference,kim2018semi}.
This motivates a basic question:
\emph{Can a trained AE itself reveal information useful for further improving
reconstruction?}

We investigate this question by repeatedly applying a frozen AE to its own reconstruction.
This produces transient image- and latent-space trajectories, which we call
\emph{Self-Reconstruction Dynamics (SRD)}.
Unlike optimization trajectories designed to approach the original image, repeated self-reconstruction in the AEs studied here progressively degrades fidelity.
Nevertheless, the resulting trajectory contains useful sample-specific information for correcting the initial reconstruction.
Rather than using later iterates as improved outputs, we therefore treat the trajectory itself as a sample-specific signal produced by the frozen AE.

Based on this observation, we propose
\emph{SRD-guided Reconstruction Refinement (SRD-RR)}, which uses a short SRD to predict a decoder-compatible latent correction while keeping the encoder and decoder frozen and requiring no per-sample test-time optimization.
Trajectory removal and cross-sample shuffling show that the gain depends on the correspondence between each sample and its own SRD, rather than only on the initial latent representation or generic AE drift.
We also introduce an $x$-free variant trained with an SRD-derived pseudo-target;
once the self-reconstruction quantities are generated, refinement training and inference require no direct access to the original image $x$.
To quantify improvement relative to the remaining capability of the fixed decoder, we introduce an \emph{MSE recovery ratio (MSE-recov)} based on an empirical decoder-optimized reference obtained by per-sample latent optimization.
It measures the fraction of the empirically recoverable MSE gap closed by a refinement method.

Across six image datasets, nonlinear SRD-conditioned refinement consistently improves reconstruction and outperforms both fixed and trained linear latent extrapolation.
Most useful information is captured by the first few trajectory transitions, and the improvement remains consistent across random seeds and AE architectures.
An additional experiment with a pretrained DINOv2-based RAE \citep{zheng2026diffusion} shows that SRD remains informative despite substantially different latent dynamics, while the refinement objective strongly affects the balance between pixel fidelity and perceptual preservation.

\textbf{Contributions.}
Our contributions are:
(1) we identify SRD as a sample-specific reconstruction signal arising
naturally from repeated application of a frozen AE, even though repeated
self-reconstruction degrades fidelity;
(2) we propose SRD-RR, including an $x$-free variant, to exploit a short SRD
for decoder-compatible latent correction without per-sample test-time
optimization; and
(3) we introduce MSE-recov for decoder-relative evaluation and systematically analyze the informativeness, sample specificity, geometry, and robustness of SRD across AEs.

\section{Related Work}
    %\section{Related Work}
\vspace{-5pt}
\paragraph{Iterated autoencoding and AE dynamics.}
Iterative or repeated application of autoencoder-based mappings has been studied in several contexts, including generative sampling \citep{bengio2013generalized,creswell2016improving}, denoising \citep{bengio2013generalized}, latent dynamics \citep{nagano2020collective}, associative memory \citep{radhakrishnan2020overparameterized}, and restoration \citep{lee2025restoration}.
Regularized AEs have also been shown to induce reconstruction vector fields related to local data structure \citep{alain2014regularized}.
More recent approaches such as Epsilon-VAE \citep{zhao2025epsilon} and Cold Diffusion \citep{bansal2023cold} use iterative processes explicitly designed for reconstruction or inversion.
SRD-RR differs in that it does not use the iteration itself as the improved
output; instead, it exploits the transient trajectory naturally produced by a
frozen AE as conditioning information for a separate latent correction.

\vspace{-7pt}
\paragraph{Latent inversion and refinement.}
Latent inversion improves reconstruction under a fixed generator or decoder,
commonly through per-image latent optimization
\citep{zhu2016generative,xia2022gan}.
Similar per-instance latent optimization has been studied in neural image
compression \citep{campos2019content,yang2020improving}, while iterative and
semi-amortized inference refine initial encoder predictions in VAEs
\citep{cremer2018inference,marino2018iterative,kim2018semi}.
% Learned iterative inversion methods such as ReStyle predict successive latent
% corrections \citep{alaluf2021restyle}, and Hong et al.\ proposed gradient-free
% decoder inversion using fixed-point and inertial iterations
% \citep{hong2024gradient}.
Learned iterative inversion methods such as ReStyle predict successive latent
corrections \citep{alaluf2021restyle}, while gradient-free decoder inversion
has been explored using fixed-point and inertial iterations
\citep{hong2024gradient}.

\vspace{-7pt}
\paragraph{Positioning relative to prior work.}
The key distinction of SRD-RR is the source of the refinement information.
Existing inversion and refinement methods generate updates as part of a target-directed procedure.
In SRD-RR, the trajectory is generated before any refinement by simply reapplying the frozen AE to its own reconstruction; it is not designed to approach the target and may instead move away from it.
SRD-RR interprets this naturally occurring trajectory as sample-specific side information and predicts a separate latent correction without per-sample optimization at inference time.
To our knowledge, prior work has not used such a naturally generated self-reconstruction trajectory for this purpose.
\section{SRD-guided Reconstruction Refinement (SRD-RR) Framework}
    %\section{SRD-Guided Reconstruction Refinement (SRD-RR) Framework}
\label{sec:method}

% 3.1
\vspace{-5pt}
\subsection{Problem Formulation and Self-Reconstruction Dynamics}
\label{sec:SRD}
\vspace{-5pt}

Let $E$ and $D$ denote the encoder and decoder, respectively, and define
$\mathrm{AE}=D\circ E$. For an input image $x$, repeated application of the
same AE gives
\[
y_1=\mathrm{AE}(x), \qquad
y_{k+1}=\mathrm{AE}(y_k), \quad k\geq1,
\]
with latent representations
\[
z_0=E(x), \qquad
z_k=E(y_k), \quad k\geq1.
\]
Thus, $y_1=D(z_0)$ and $y_{k+1}=D(z_k)$. We define the reverse-step
vectors
\[
dy_k=y_k-y_{k+1}, \qquad
dz_k=z_k-z_{k+1},
\]
which point opposite to the direction of repeated autoencoding.

We refer to the sequences
$\{y_k\}_{k\geq1}$ and $\{z_k\}_{k\geq1}$ as the
\emph{Self-Reconstruction Dynamics (SRD)}.
SRD-guided Reconstruction Refinement (SRD-RR) uses information from this
trajectory to obtain a reconstruction $\hat{x}$ that improves upon the
one-shot reconstruction $y_1$.
%
% We consider post-hoc refinement with access to $y_1$ and the frozen AE, but not to the original image $x$ or its initial encoder output $z_0$.
% %
% Because $z_0=E(x)$ and $dz_0=z_0-z_1$ directly depend on the original
% input, they are unavailable to a refinement procedure operating only from
% $y_1$ onward. Unless otherwise stated, SRD-RR therefore uses the observable
% trajectory information
% \[
% z_1,\;dz_1,\;dz_2,\ldots,
% \]
% which can be generated solely by repeatedly applying the fixed AE to $y_1$.

% We denote the number of observed SRD transitions by $N_{\mathrm{tr}}$.
% Thus, a trajectory generated up to $y_K$ provides  $N_{\mathrm{tr}}=K-1$ transitions, $\{dz_1,\ldots,dz_{K-1}\}$.
%
We consider post-hoc refinement from $y_1$.
At refinement inference, only $y_1$ and the frozen AE are available; $x$ and $z_0$ are not retained.
Thus, SRD-RR uses the observable latent trajectory
\[
z_1,\;dz_1,\;dz_2,\ldots,
\]
generated solely by repeatedly applying the fixed AE to $y_1$.

% 3.2
\vspace{-5pt}
\subsection{Empirical Decoder-optimized Reference and Evaluation}
\label{sec:decoder_bound}
\vspace{-5pt}

To quantify the reconstruction headroom of the fixed decoder, we optimize a latent representation for each sample while keeping $D$ fixed:
\begin{equation}
z^\ast
=
\arg\min_z \|D(z)-x\|_2^2,
\label{eq:zstar}
\end{equation}
initialized from $z_0=E(x)$. This follows standard latent-optimization
practice in image inversion
\citep{zhu2016generative,xia2022gan}.

We denote $y^\ast=D(z^\ast)$ as the \emph{empirical decoder-optimized reference}.
Because ~\eqref{eq:zstar} is solved numerically and does not guarantee a global optimum, this is an empirical reconstruction reference rather than a theoretical bound.

For a refined reconstruction $\hat{x}$, we define the
\emph{MSE recovery ratio} (MSE-recov) as
\begin{equation}
\operatorname{MSE\text{-}recov}(\hat{x})
=
\frac{
\operatorname{MSE}(y_1,x)-\operatorname{MSE}(\hat{x},x)
}{
\operatorname{MSE}(y_1,x)-\operatorname{MSE}(y^\ast,x)
},
\label{eq:mse_recovery}
\end{equation}
% where
% $\operatorname{MSE}(a,x)=N^{-1}\|a-x\|_2^2$.
A value of $0$ corresponds to the baseline reconstruction $y_1$, $1$ to the empirical decoder-optimized reference $y^\ast$, and a negative value indicates degradation from $y_1$.
Values above 1 are possible for methods not constrained to the output range of the fixed decoder, as they may achieve lower MSE than the decoder-optimized reference $y^\ast=D(z^\ast)$.
In our experiments, this occurs only for the image-space M3 on two datasets.
MSE-recov is computed independently for each sample before averaging over the dataset.

For latent-space geometry analysis, we additionally define the
\emph{empirical correction vector}
\begin{equation}
v^\ast=z^\ast-z_1,
\label{eq:vstar}
\end{equation}
which provides a sample-specific reference direction for comparison with the
SRD reverse-step vectors $dz_k$.

% 3.3
\vspace{-5pt}
\subsection{\texorpdfstring{Model Variants and $x$-Free Refinement}{Model Variants and x-Free Refinement}}

\label{sec:models}
% Table 1
\begin{table}[t]
\centering
\vspace{-4pt}
\caption{Summary of evaluated models, supervision, and $x$-free status.}
\label{tab:model_summary}
%\vspace{-2pt}

\renewcommand{\arraystretch}{1.2}
\resizebox{\textwidth}{!}{
\begin{tabular}{l l l c c}
\hline
\textbf{Model} &
\textbf{Description} &
\textbf{Definition} &
\textbf{Target} &
\textbf{$x$-free} \\
\hline

\multicolumn{5}{l}{\textit{\large Baseline and image-space SRD models}} \\

M1 &
Baseline AE &
$y_1$ &
-- &
-- \\

M2 &
Repeated AE &
$y_K$ &
-- &
-- \\

M3 &
Trained Nonlinear Image Corrector &
$\tilde{x}=y_1+\Delta x,\quad
\Delta x=M_3(y_1,dy_1,\ldots,dy_{K-1})$ &
{\large $x$} &
-- \\

M3-AE &
Corrector + AE &
$\mathrm{AE}(M3)$ &
{\large $x$} &
-- \\

\bf{M4} &
Linear Image Extrapolation &
$\tilde{x}_{\mathrm{lin}}
=y_1+\alpha(y_1-y_2)$ &
-- &
\checkmark \\

\bf{M4-AE} &
Linear Image Extrapolation + AE &
$\mathrm{AE}(M4)$ &
-- &
\checkmark \\

\hline
\multicolumn{5}{l}{\textit{\large Linear latent-space SRD models}} \\

\bf{M5} &
Linear Latent Extrapolation &
$D(\tilde{z}_{\mathrm{lin}}),\quad
\tilde{z}_{\mathrm{lin}}=z_1+\beta_1 dz_1$ &
-- &
\checkmark \\

\bf{M5-AE} &
Linear Latent Extrapolation + AE &
$\mathrm{AE}(M5)$ &
-- &
\checkmark \\

M5-T &
Trained Linear Trajectory Correction &
$D(\tilde{z}_{\mathrm{lin}}^{*}),\quad
\tilde{z}_{\mathrm{lin}}^{*}
=z_1+\sum_{k=1}^{K-1}\beta_k^{*}dz_k$ &
{\large $x$} &
-- \\

\hline
\multicolumn{5}{l}{\textit{\large Nonlinear latent-space SRD models; ~  $D(z_1+ \gamma \Delta z)$}} \\

M6-LO &
Latent-Only Corrector &
$\Delta z=M_{6\mathrm{LO}}(z_1)$ &
{\large $x$} &
-- \\

M6 &
SRD-RR &
$\Delta z=M_6(z_1,dz_1,\ldots,dz_{K-1})$ &
{\large $x$} &
-- \\

M6-Lobj &
Latent-Objective SRD-RR &
% $\begin{aligned}
% D(z_1+\Delta z),\\
% \Delta z&=M_{6\mathrm{LAT}}(z_1,dz_1,\ldots,dz_{K-1})
% %\Delta z&=M_{6LAT}(z_1,dz_1,\ldots,dz_{K-1})
% \end{aligned}$ &
$\Delta z=M_{6\mathrm{LAT}}(z_1,dz_1,\ldots,dz_{K-1})$&
{\large $z^{*}$} &
-- \\

\bf{M6-xFree} &
$x$-free SRD-RR &
% $D(z_1+\Delta z),\quad
% \Delta z=M_{6\mathrm{D}}(z_1,dz_1,\ldots,dz_{K-1})$ &
$\Delta z=M_{6\mathrm{DEP}}(z_1,dz_1,\ldots,dz_{K-1})$ &
{\large $\tilde{x}_{\mathrm{lin}}$} &
\checkmark \\

M6-Shuf &
Shuffled-Trajectory SRD-RR $\star$ &
% $\begin{aligned}
% D(z_1^{(i)}+\Delta z^{(i)}),\\
% \Delta z^{(i)}
% &=M_6\bigl(
% z_1^{(i)},
% dz_1^{(\pi(i))},\ldots,
% dz_{K-1}^{(\pi(i))}
% \bigr)
% \end{aligned}$ &
%$\Delta z^{(i)}=M_{6\mathrm{SHU}}
$\Delta z^{(i)}=M_6(z_1^{(i)},dz_1^{(\pi(i))},\ldots,dz_{K-1}^{(\pi(i))})$ &
{\large $x$} &
-- \\

M7 &
SRD-RR + $\tilde{z}_0 ~ \dagger$ &
%\begin{aligned}
% D(z_1+\Delta z),\\
% \Delta z
% &=M_7(z_1,dz_1,\ldots,dz_{K-1},\tilde{z}_0)
% \end{aligned}$ &
$\Delta z=M_7(z_1,dz_1,\ldots,dz_{K-1},\tilde{z}_0)$&
{\large $x$} &
-- \\

\bf{M8} &
SRD-RR + $\tilde{z}_{0,\mathrm{lin}} ~ \ddagger$ &
% $\begin{aligned}
% D(z_1+\Delta z),\\
% \Delta z
% &=M_8(z_1,dz_1,\ldots,dz_{K-1},
% \tilde{z}_{0,\mathrm{lin}})
% \end{aligned}$ &
$\Delta z=M_8(z_1,dz_1,\ldots,dz_{K-1},\tilde{z}_{0,\mathrm{lin}})$&
{\large $\tilde{x}_{\mathrm{lin}}$} &
\checkmark \\

\hline

\multicolumn{2}{l}{\textit{\large Empirical Decoder-optimized  Reference}} &
$D(z^{*})$ &
{\large $x$} &
-- \\

\hline
\end{tabular}
}

\vspace{2pt}

{\scriptsize
\raggedright
% $^\star$: $D(z_1^{(i)}+\Delta z^{(i)})$, trajectories $dz_k$ is taken from another sample under permutation $\pi$.\\
$^\star$: Test-time shuffling uses the trained M6 predictor without retraining.\\
$^\dagger$:
$\tilde{z}_0=E(\tilde{x})$, where $\tilde{x}$ is the output of M3.\\
$^\ddagger$:
$\tilde{z}_{0,\mathrm{lin}}=E(\tilde{x}_{\mathrm{lin}})$, where
$\tilde{x}_{\mathrm{lin}}$ is the output of M4 ($x$-free).
\par}

\end{table}
\vspace{-5pt}
Table~\ref{tab:model_summary} summarizes all evaluated models.
SRD-RR predicts a latent correction from the observable trajectory as
\[
\hat{z}
=
z_1+\gamma\Delta z_\theta
(z_1,dz_1,\ldots,dz_{N_{\mathrm{tr}}}),
\qquad
\hat{x}=D(\hat{z}).
\]
Here, $N_{\mathrm{tr}}=K-1$ denotes the number of observed SRD transitions, and $\gamma$ is a learnable scalar.
M6 minimizes $\|\hat{x}-x\|_2^2$, whereas M6-xFree replaces $x$
with the SRD-derived pseudo-target
$\tilde{x}_{\mathrm{lin}}=2y_1-y_2$, i.e., the M4 output with $\alpha=1$.
All latent-space correctors are anchored at $z_1=E(y_1)$, so zero correction yields $D(z_1)=y_2$ rather than $y_1$. Unlike M3 and M4, which are anchored directly at $y_1$, they must first recover the $y_1\!\to y_2$ degradation before improving upon the one-shot baseline.

We call a refinement method $x$-free if, once $y_1$ has been obtained, neither its training nor refinement inference requires direct access to $x$ or any ground-truth-dependent quantity.
Under this definition, M4, M4-AE, M5, M5-AE, M6-xFree, and M8 are $x$-free.

\section{Experiments}
    %\section{Experiments}
\label{sec:experiments}

%4.1
\vspace{-5pt}
\subsection{Common Experimental Setup}
\label{sec:common_setup}
\vspace{-5pt}
\paragraph{Datasets.}
We evaluate SRD-RR on CIFAR-10 \citep{krizhevsky2009learning},
SVHN \citep{netzer2011reading}, FashionMNIST \citep{xiao2017fashion},
MNIST \citep{lecun1998gradient},
ImageNet100 \citep{deng2009imagenet,russakovsky2015imagenet},
and CelebA \citep{liu2015deep}.
We abbreviate FashionMNIST and ImageNet100 as FMNIST and ImageNet,
respectively.

Dataset splits are summarized in Table~\ref{tab:dataset_split}.
For the first four datasets, the original training set is split 9:1 for training and validation, with the official test set used for evaluation.
For ImageNet, the original training set is split similarly, with 5,000 fixed samples from the original validation set used for testing.
For CelebA, we split the original training set 9:1 and use 5,000 fixed samples from the official test set.
For ImageNet and CelebA, the nonlinear trajectory predictors (M6 family, M7, and M8) use fixed subsets of 20,000 training and 2,000 validation samples because of the storage cost of multi-step latent trajectories.
The empirical latent $z^\ast$ is used only for evaluation and analysis, except in M6-Lobj, where $z^\ast$ computed from training samples serves as the regression target.
Image sizes and pre-processing are described in Appendix \ref{app:pre-processing}.  
%
% \begin{table}[t]
% \centering
% \caption{Dataset splits used in the experiments.}
% \small
% \label{tab:dataset_split}
% \begin{tabular}{lrrr}
% \toprule
% \textbf{Dataset} & \textbf{Train} & \textbf{Val} & \textbf{Test} \\
% \midrule
% CIFAR-10      & 45,000  & 5,000  & 10,000 \\
% SVHN          & 65,932  & 7,325  & 26,032 \\
% FashionMNIST  & 54,000  & 6,000  & 10,000 \\
% MNIST         & 54,000  & 6,000  & 10,000 \\
% ImageNet100   & 114,021 & 12,668 & 5,000 \\
% CelebA        & 146,493 & 16,277 & 5,000 \\
% \bottomrule
% \end{tabular}
% \end{table}

\begin{table}[t]
\centering
\caption{Dataset splits used in the experiments.}
\small
\label{tab:dataset_split}
\begin{tabular}{lrrrrrr}
\toprule
\textbf{Split}
& \textbf{CIFAR-10}
& \textbf{SVHN}
& \textbf{FashionMNIST}
& \textbf{MNIST}
& \textbf{ImageNet100}
& \textbf{CelebA} \\
\midrule
Train
& 45,000
& 65,932
& 54,000
& 54,000
& 114,021
& 146,493 \\

Val
& 5,000
& 7,325
& 6,000
& 6,000
& 12,668
& 16,277 \\

Test
& 10,000
& 26,032
& 10,000
& 10,000
& 5,000
& 5,000 \\
\bottomrule
\end{tabular}
\end{table}
\vspace{-5pt}
\paragraph{Evaluation.}
We use MSE-recov and PSNR as the primary reconstruction metrics.
Both are computed independently for each test sample and then averaged within each dataset.
Cross-dataset results are macro-averaged over the six datasets.
Inference latency is measured on 500 ImageNet images with batch size 1 on an NVIDIA RTX 6000 Pro GPU, including all computation from $x$ to the final output but excluding model loading.
Details of preprocessing, model architecture, training, and normalization are provided in
Appendix~\ref{app:implementation}.

%4.2
\vspace{-5pt}
\subsection{Experiment 1: Image-Space Correction}
\label{sec:exp_image}
\vspace{-5pt}
We first test whether image-space SRD contains reconstruction-relevant information.
M1 and M2 denote one-shot and repeated AE reconstruction, respectively.
M3 learns a nonlinear residual correction $\Delta x$ from
$(y_1,dy_1,\ldots,dy_{K-1})$, whereas M4 performs linear image-space extrapolation.
M4 is related in form to unsharp masking \citep{polesel2000image}, although
its residual is induced by self-reconstruction rather than an explicit blurring filter.
We fix $\alpha=1.0$, corresponding to direct linear extrapolation of the first self-reconstruction step.
A validation-set sweep confirms that this choice performs well across all datasets (Appendix Fig.~\ref{fig:alpha_beta_sweep}).
M3-AE and M4-AE pass the corrected images through the AE once more to test whether the improvement survives re-encoding.

%4.3
\vspace{-5pt}
\subsection{Experiment 2: Linear Latent-Space Correction}
\label{sec:exp_linear_latent}
\vspace{-5pt}
We evaluate fixed linear latent extrapolation M5 and trained linear trajectory correction M5-T (Table~1) to test whether latent SRD provides a useful linear correction direction.
Validation sweeps consistently favor $\beta\approx2.0$ across all six datasets; we therefore use $\beta=2.0$ for final test evaluation.
Notably, the independently trained M5-T also learns first-step coefficients close to 2.0.
Because M5 is training-free, computationally efficient, decoder-compatible, and $x$-free, we additionally study repeated application of M5 in Appendix~\ref{app:results}.

%4.4
\vspace{-5pt}
\subsection{Experiment 3: Geometry of SRD}
\label{sec:exp_geometry}
\vspace{-5pt}
We analyze the local coherence, curvature, and reconstruction relevance of latent SRD using
$\cos(dz_k,dz_{k+1})$, $\cos(dz_0,dz_k)$, and
$\cos(v^\ast,dz_k)$.
These quantify successive directional consistency, deviation from the initial trajectory direction, and alignment with the empirical correction direction, respectively.
The quantities $dz_0$ and $v^\ast$ are used only for analysis.

% 4.5
\vspace{-5pt}
\subsection{Experiment 4: Nonlinear SRD-RR and Robustness}
\label{sec:exp_nonlinear}
\vspace{-5pt}

We compare nonlinear SRD-RR M6 with several variants and ablations.
M6-LO serves as a trajectory-free baseline that uses only $z_1$, allowing us to isolate the contribution of the SRD trajectory.
We also evaluate the latent-objective M6-Lobj, $x$-free M6-xFree, shuffled-trajectory M6-Shuf, and auxiliary-guide variants M7 and M8
(Table~\ref{tab:model_summary}), using $N_{\mathrm{tr}}\in\{1,2,4\}$.
For M6-Shuf, we keep the trained predictor and each sample's $z_1$ fixed and replace its trajectory with that of another sample at evaluation time.

We additionally evaluate robustness to training randomness and AE architecture.
Experiments on CIFAR-10 and ImageNet are repeated with five random seeds, and three AE configurations with different depths and channel capacities are evaluated.

% 4.6
\vspace{-5pt}
\subsection{Experiment 5: Extension to a Pretrained Representation Autoencoder}
\label{sec:exp_RAE}
\vspace{-5pt}
To examine SRD in a substantially different autoencoder, we use a pretrained DINOv2-based representation autoencoder (RAE) to address two questions: (1) whether SRD remains informative under qualitatively different latent dynamics, and (2) how the refinement objective affects pixel and perceptual reconstruction quality.
Both RAE components are kept frozen in our experiments.
The SRD-RR predictor uses the same predictor backbone as in the ImageNet experiments, with RAE-specific settings described in Appendix~\ref{app:RAE}.
We use 20,000/2,000/5,000 images for predictor training, validation, and
testing, respectively, and evaluate M1, repeated reconstruction (M2), M4, M5, M6-LO, M6 ($N_{\rm tr}=1,2$), and M6-xFree ($N_{\rm tr}=1,2$).
Here, M6-LO serves as the matched trajectory-free baseline for isolating the contribution of SRD conditioning.

Evaluation uses MSE, PSNR, SSIM \citep{wang2004image},
LPIPS \citep{zhang2018unreasonable}, and rFID-5k, computed as the
Fr\'echet Inception Distance (FID) \citep{heusel2017gans} between the original and reconstructed image distributions on the fixed 5,000-image test set.
Additional RAE-specific implementation and evaluation details are provided in
Appendix~\ref{app:RAE}.
\section{Results}
    %\section{Results}
\label{sec:results}

%5.1
\vspace{-5pt}
\subsection{Image-Space Correction}
\label{sec:result_image}
\begin{table*}[t]
\centering
\caption{MSE recovery ratio (\%) of image-space SRD methods.}
%The one-shot reconstruction (M1) is defined as 0\%, and the empirical oracle
%reconstruction $D(z^\ast)$ as 100\%.}
\label{tab:image_space_results}
\small
\setlength{\tabcolsep}{3pt}
\begin{tabular}{llrrrrrr}
\toprule
\textbf{Model} &
\textbf{Description} &
\textbf{CIFAR-10} &
\textbf{SVHN} &
\textbf{FMNIST} &
\textbf{MNIST} &
\textbf{ImageNet} &
\textbf{CelebA} \\
\midrule

M1
& $y_1$
%& 0.00 & 0.00 & 0.00 & 0.00 & 0.00 & 0.00 \\
&\multicolumn{6}{c}{0}\\
\midrule
% M2 ($K=2$)
% & $y_2$
% & -131.35 & -179.44 & -143.94 & -155.45 & -161.00 & -157.26 \\

% M2 ($K=3$)
% & $y_3$
% & -311.27 & -431.78 & -337.00 & -350.96 & -392.95 & -370.97 \\

% M2 ($K=5$)
% & $y_5$
% & -802.66 & -1125.70 & -864.78 & -819.55 & -1065.15 & -959.68 \\

\multirow{3}{*}{M2}
& $y_2$ \;
%& -131.35 & -179.44 & -143.94 & -155.45 & -161.00 & -157.26 \\
&-143.02	&-197.19	&-157.09	&-158.24	&-173.95	&-172.62\\

& $y_3$ \;
%& -311.27 & -431.78 & -337.00 & -350.96 & -392.95 & -370.97 \\
&-346.00 &	-488.55	& -371.01 &	-360.08	& -429.96 &	-412.22 \\

& $y_5$ \;
%& -802.66 & -1125.70 & -864.78 & -819.55 & -1065.15 & -959.68 \\
&-917.10	&-1324.69		&-961.63	&-849.91	&-1184.02	&-1086.96\\

\midrule

M3
& $\tilde{x}$
%& 115.37 & 103.74 & 83.36 & 95.51 & 53.11 & 95.97 \\
&116.87	&103.86	&87.09	&96.20	&51.54	&98.73 \\

M3-AE
& AE(M3)
%& -9.02 & -7.48 & -5.98 & -5.77 & 8.71 & -13.71 \\
&-8.93	&-6.16	&-5.38	&-5.37	&10.54	&-14.03 \\

M4 ($\alpha=1.0$)
& $\tilde{x}_{\mathrm{lin}}$
%& 50.20 & 64.02 & 57.51 & 62.82 & 59.75 & 58.87 \\
&52.72	&68.33	&61.88	&63.75	&63.36	&63.09 \\

M4-AE
& AE(M4)
%& -16.19 & -21.96 & -20.68 & -24.36 & -15.77 & -23.39 \\
&-14.52	&-16.64	&-21.05	&-24.17	&-14.96	&-23.05\\

\midrule

{\footnotesize D-optimized Reference}
& $y^\ast = D(z^\ast)$
%& 100.00 & 100.00 & 100.00 & 100.00 & 100.00 & 100.00 \\
&\multicolumn{6}{c}{100}\\
\bottomrule
\end{tabular}
\end{table*}
\vspace{-5pt}
Table~\ref{tab:image_space_results} summarizes the image-space results.
Repeated autoencoding (M2) progressively degrades reconstruction as iteration increases, whereas both learned (M3) and linear (M4) correction substantially improve the baseline reconstruction.
M3 achieves an average MSE-recov of $92.4\%$ and exceeds the empirical decoder-optimized reference on two datasets.
The auxiliary M4 sweep peaks near $\alpha=1.0$ across datasets
(Appendix Fig.~\ref{fig:alpha_beta_sweep}), supporting our fixed choice $\alpha=1.0$.
However, these gains largely vanish after re-encoding: M3-AE and M4-AE perform at or below the baseline in most cases.
Thus, strong image-space improvement does not necessarily correspond to a decoder-compatible latent representation.

% 5.2
\vspace{-5pt}
\subsection{Linear Latent-Space Correction}
\label{sec:result_linear}
\begin{table*}[t]
\centering
\caption{MSE recovery ratio (\%) of linear latent-space correctors and estimated $\beta_1^\ast$.}
\label{tab:linear_latent_corrector}
\small
\setlength{\tabcolsep}{6pt}
\begin{tabular}{lcrrrrrr}
\toprule
\textbf{Model} &
$\mathbf{N_{\mathrm{tr}}}$ &
\textbf{CIFAR-10} &
\textbf{SVHN} &
\textbf{FMNIST} &
\textbf{MNIST} &
\textbf{ImageNet} &
\textbf{CelebA} \\
\midrule

M5 ($\beta=2.0$)
& 1
%& 20.90 & 18.22 & 14.05 & 11.86 & 28.63 & 20.16 \\
&24.97	&34.58	&17.59	&12.55	&33.09	&24.96 \\

M5-AE
& 1
%& -43.65 & -66.36 & -52.67 & -56.09 & -39.42 & -65.32 \\
&-40.84	&-53.22	&-54.55	&-55.69	&-38.46	&-66.18 \\

\midrule

\multirow{4}{*}{M5-T}
& 1
%& 21.22 & 19.52 & 14.71 & 12.86 & 28.39 & 16.39 \\
&24.99 	&33.89 	&18.03 	&13.66 	&32.15 	&19.42 \\

& 2
%& 29.62 & 31.90 & 25.00 & 27.97 & 23.73 & 3.28 \\
&32.48 	&43.79 	&29.04 	&28.96 	&27.36 	&4.64 \\

& 3
%& 29.62 & 32.38 & 25.65 & 30.55 & 20.34 & 27.05 \\
&32.52 	&43.36 	&29.68 	&31.51 	&24.36 	&31.10\\

& 4
%& 29.62 & 33.33 & 25.65 & 30.55 & 13.56 & 23.77 \\
&32.53 	&44.75 	&29.73 	&31.59 	&16.89 	&28.23\\

\midrule

\multicolumn{2}{l}{\footnotesize M5-T: estimated $\beta_1^\ast$} {\scriptsize $(N_{tr}=1)$}
& 1.983
& 1.817
& 1.910
& 1.868
& 2.077
& 2.418 \\

\bottomrule
\end{tabular}
\end{table*}

\vspace{-5pt}
Table~\ref{tab:linear_latent_corrector} shows the linear latent-space results.
M5 improves reconstruction on all six datasets.
The validation sweep consistently favors $\beta\approx2.0$ (Appendix Fig.~\ref{fig:alpha_beta_sweep}), consistent with the first coefficient independently learned by M5-T 
(Table \ref{tab:linear_latent_corrector}).
Additional transitions improve M5-T on the four smaller datasets, with diminishing gains, but produce dataset-dependent degradation or fluctuations on ImageNet and CelebA.
M5-AE again falls below the baseline, indicating that the encoder does not preserve the refined latent after decoding and re-encoding.
Repeated application of M5 provides no additional benefit beyond the first correction (Appendix Fig.~\ref{fig:iterativeM5}), consistent with the changing direction of the latent SRD.
These results motivate a nonlinear correction that explicitly exploits the short SRD.

% 5.3
\vspace{-5pt}
\subsection{Geometry of SRD}
\label{sec:result_geometry}
% \begin{figure}[t]
%     \centering
%     \includegraphics[width=\columnwidth]{images/sec3_trajectory_main_ver4.png}
%     \caption{Geometric properties of the latent Self-reconstruction Dynamics (SRD).}
%     \label{fig:SRD_trajectory}
% \end{figure}

\begin{figure}[t]
    \centering

    \begin{subfigure}[t]{0.32\textwidth}
        \centering
        \includegraphics[width=\linewidth]{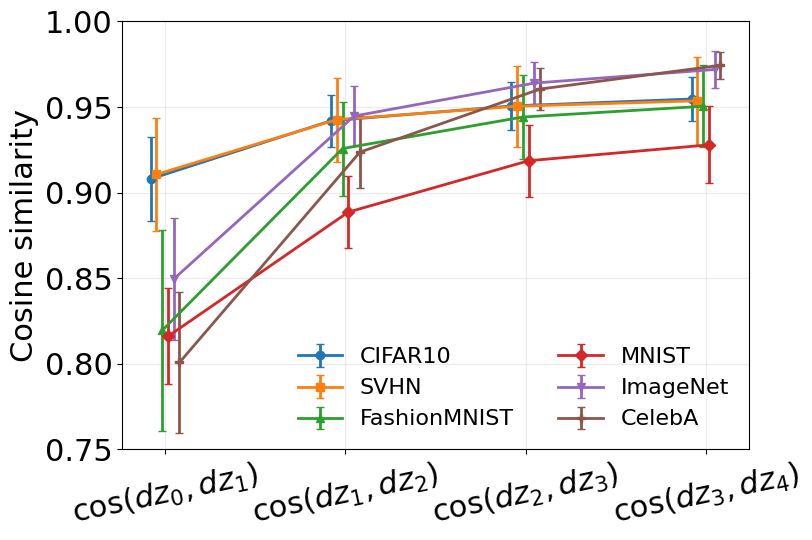}
        \caption{$\cos(dz_k, dz_{k+1})$}
        \label{fig:SRD_trajectory_a}
    \end{subfigure}
    \hfill
    \begin{subfigure}[t]{0.32\textwidth}
        \centering
        \includegraphics[width=\linewidth]{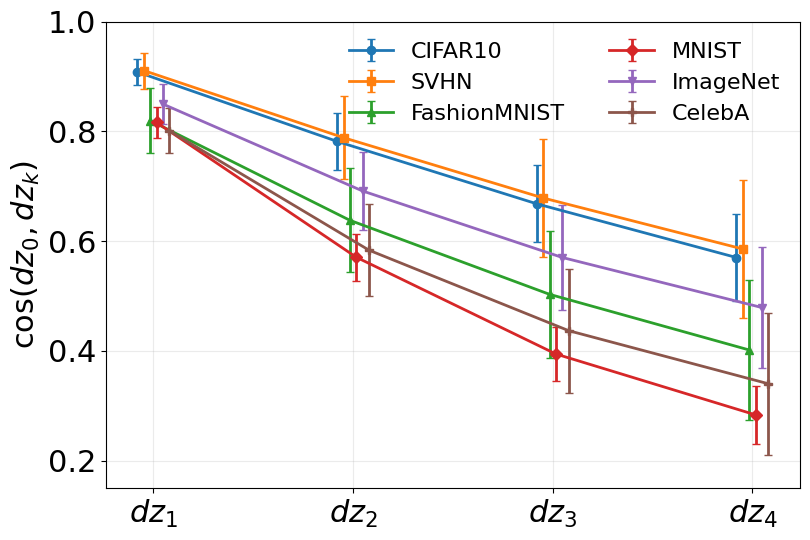}
        \caption{$\cos(dz_0, dz_k)$}
        \label{fig:SRD_trajectory_b}
    \end{subfigure}
    \hfill
    \begin{subfigure}[t]{0.32\textwidth}
        \centering
        \includegraphics[width=\linewidth]{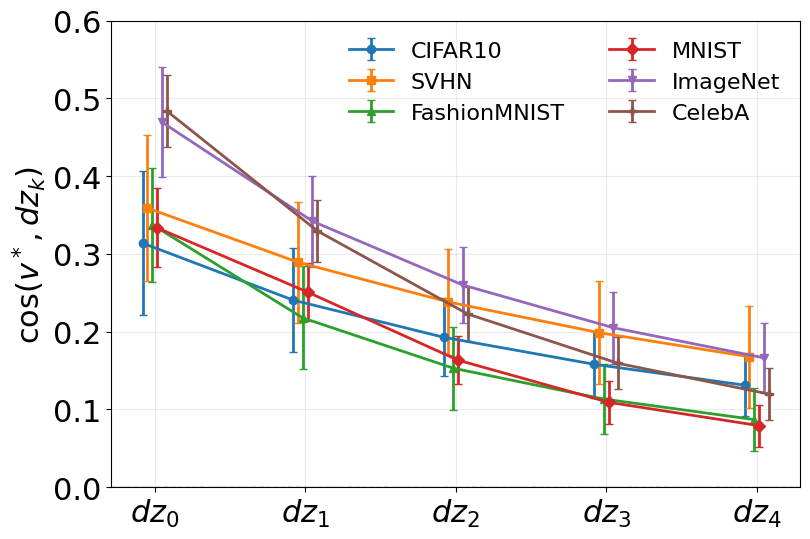}
        \caption{$\cos(v^\ast, dz_k)$}
        \label{fig:SRD_trajectory_c}
    \end{subfigure}

    \caption{Geometric properties of the latent Self-reconstruction Dynamics (SRD). Error bars indicate standard deviations within each dataset.}
    \label{fig:SRD_trajectory}
\end{figure}
\vspace{-5pt}
Figure~\ref{fig:SRD_trajectory} characterizes the latent-space geometry of SRD.
Successive reverse-step vectors have high cosine similarity, indicating strong local directional coherence, while their alignment with the initial direction decreases over time, revealing a curved rather than globally linear trajectory (Appendix Fig.~\ref{fig:trajectory_all}).
The observable $dz_k$ also exhibit moderate positive alignment with the empirical correction vector $v^\ast$, particularly at early steps.
Thus, early SRD contains information related to the reconstruction-improving
direction without directly coinciding with it.

% 5.4
\vspace{-5pt}
\subsection{Nonlinear SRD-RR and Sample Specificity}
\label{sec:result_nonlinear}
\begin{figure*}[t]
    \centering

    \begin{subfigure}[t]{0.48\textwidth}
        \centering
        \includegraphics[width=\linewidth]{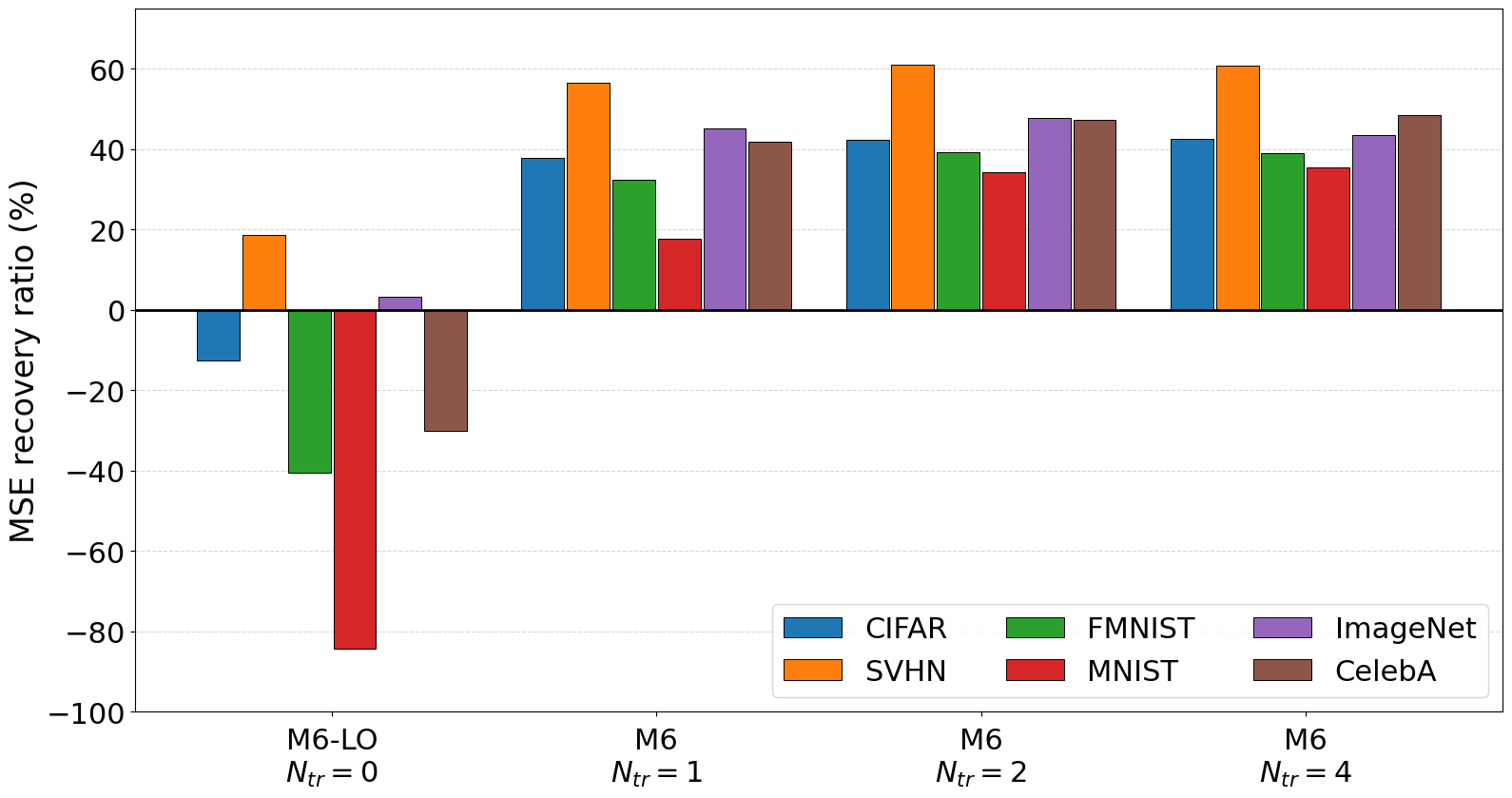}
        \caption{Trajectory contribution}
        \label{fig:M6models}
    \end{subfigure}
    \hspace{5mm}
    \begin{subfigure}[t]{0.30\textwidth}
        \centering
        \includegraphics[width=\linewidth]{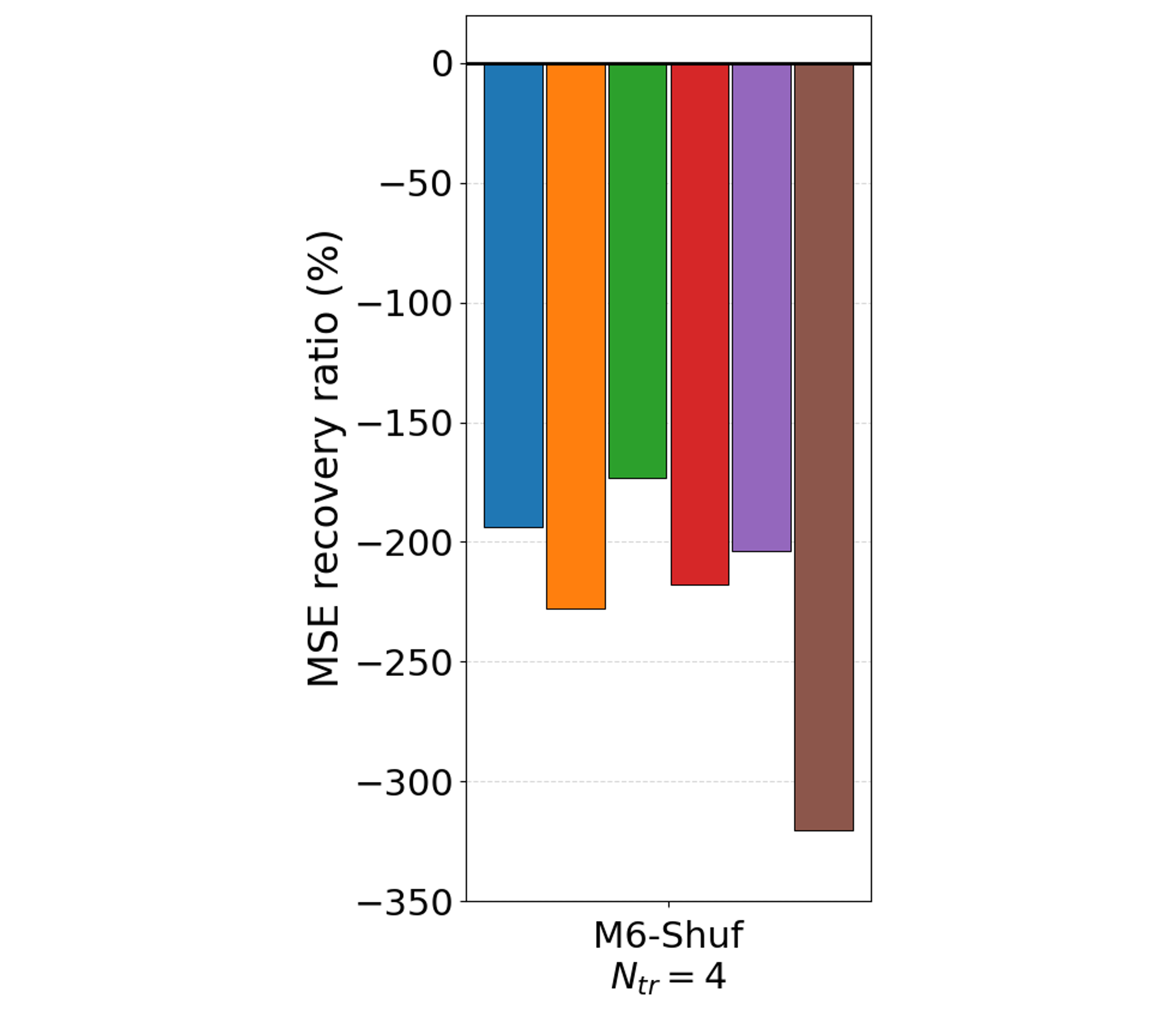}
        \caption{Trajectory shuffling}
        \label{fig:M6-Shuffle}
    \end{subfigure}

    \caption{Effect of trajectory information and sample correspondence on nonlinear SRD-RR models.}
    \label{fig:M6family}
\end{figure*}

% \begin{figure*}[t]
%     \centering

%     \begin{subfigure}[t]{0.48\textwidth}
%         \centering
%         \includegraphics[width=\linewidth]{images/M6family_compare.png}
%         \caption{}
%         \label{fig:M6models}
%     \end{subfigure}
%     \hspace{0.01\textwidth}
%     \begin{subfigure}[t]{0.15\textwidth}
%         \centering
%         \includegraphics[width=\linewidth]{images/M6_shuffle_compare.png}
%         \caption{}
%         \label{fig:M6-Shuffle}
%     \end{subfigure}

%     \caption{
%     Effect of trajectory information and sample correspondence on nonlinear
%     SRD-RR models.
%     (a) Performance of M6-LO and M6 with different numbers of trajectory steps
%     $N_{\mathrm{tr}}$.
%     (b) Effect of test-time trajectory shuffling for M6 with
%     $N_{\mathrm{tr}}=4$.
%     }
%     \label{fig:M6family}
% \end{figure*}
\vspace{-5pt}
Figure~\ref{fig:M6family} shows that M6 consistently outperforms M6-LO, demonstrating that explicitly providing SRD features improves correction prediction over using $z_1$ alone.
Most of the gain is captured by the first few steps and largely saturates by $N_{\mathrm{tr}}=2$.
Cross-sample trajectory shuffling causes severe degradation on every dataset.
The benefit therefore depends on the correspondence between each sample and its own SRD, rather than on generic AE drift.

% 5.5
\vspace{-5pt}
\subsection{Overall Performance, Efficiency and Robustness}
\label{sec:result_overall}
\begin{table*}[t]
\centering
\caption{Summary of reconstruction performance, computational cost, and \(x\)-free status.}
\label{tab:overall_performance}

\small
\setlength{\tabcolsep}{2.5pt}

\begin{tabular}{lc@{\hspace{2pt}}rrrrrr|r|c}
\toprule
& &
\multicolumn{6}{c|}{\textbf{MSE-recov (\%)}}
& \textbf{Cost~}
& \\

\textbf{Model}
& $\mathbf{N_{\mathrm{tr}}}$
& \textbf{CIFAR-10}
& \textbf{SVHN}
& \textbf{FMNIST}
& \textbf{MNIST}
& \textbf{ImageNet}
& \textbf{CelebA}
& $\times$\textbf{M1}
& \textbf{$x$-free} \\
\midrule

M1
& --
& \multicolumn{6}{c|}{0}
& 1.0 & -- \\
\midrule

M3 $^\dagger$
& 1
% & 115.37 & 103.74 & 83.36 & 95.51 & 53.11 & 95.97
& 116.87 &103.86 &87.09	&96.2	&51.54	&98.73
& 8.1 & -- \\

M4 ($\alpha=1.0$) $^\dagger$
& 1
%& 50.20 & 64.02 & 57.51 & 62.82 & 59.75 & 58.87
&52.72	&68.33	&61.88	&63.75	&63.36	&63.09
& 2.0 & \checkmark \\

\midrule

M5 ($\beta=2.0$)
& 1
%& 20.90 & 18.22 & 14.05 & 11.86 & 28.63 & 20.16
&24.97	&34.58	&17.59	&12.55	&33.09	&24.96
& 2.9 & \checkmark \\

\midrule

M6-LO
& 0
%& -16.39 & 9.81 & -44.91 & -85.58 & -4.56 & -35.48
&-12.71	&18.62	&-40.63	&-84.45	&3.28	&-30.06
& 4.2 & -- \\

M6
& 1
%& 34.63 & 50.00 & 26.98 & 15.71 & 40.66 & 37.10
&37.84	&56.56	&32.34	&17.6	&45.12	&41.85
& 6.1 & -- \\

M6
& 2
%& 39.75 & 55.14 & 33.60 & 31.73 & 43.57 & 42.74
&42.42	&61.05	&39.20	&34.32	&47.72	&47.21
& 6.3 & -- \\

M6
& 4
%& 39.96 & 54.67 & 33.28 & 32.69 & 44.40 & 43.55
&42.49 	&60.68 	&39.10 	&35.41 	&43.47 	&48.47 
& 9.0 & -- \\

M6-Lobj
& 2
%& 32.17 & 42.52 & -273.34 & -17543.59 & 39.83 & -94.35
&35.23	&51.44	&-289.58	&-19294.8	&44.56	&-93.41
& 5.8 & -- \\

\bf{M6-xFree}
& \bf{2}
% & \bf{32.38} & \bf{44.39} & \bf{30.53} & \bf{30.12}
% & \bf{42.73} & \bf{35.48}
&\bf{35.99}	&\bf{52.85}	&\bf{35.69}	&\bf{31.68}	&\bf{47.24}	&\bf{40.46}
& \bf{6.3} & \checkmark \\

M7
& 2
%& 40.78 & 58.41 & 38.45 & 41.35 & 29.88 & 40.32
&43.19	&63.31	&45.06	&44.08	&30.22	&45.12
& 12.1 & -- \\

\bf{M8}
& \bf{2}
% & \bf{34.22} & \bf{44.86} & \bf{31.18} & \bf{31.09}
% & \bf{42.32} & \bf{35.48}
&\bf{37.65}	&\bf{53.11}	&\bf{36.28}	&\bf{32.79}	&\bf{47.03}	&\bf{40.16}
&\bf{6.9} & \checkmark \\

\midrule

%\multicolumn{2}{l}{\small Decoder reference} $D(z^\ast)$
\multicolumn{2}{l}{{\footnotesize D-optimized Reference $D(z^\ast)$}}
& \multicolumn{6}{c|}{100}
& 1317 & -- \\

\bottomrule
\end{tabular}

{\scriptsize
\raggedright
$^\dagger$ M3 and M4 improve reconstruction quality without enforcing decoder compatibility; their gains diminish after re-encoding through the AE (see Table~\ref{tab:image_space_results}).  Bold rows highlight the learned x-free variants.\\
\par}
\end{table*}
\vspace{-5pt}
% Table~\ref{tab:overall_performance} summarizes the principal models in terms of MSE recovery.
%
Table~\ref{tab:overall_performance} summarizes the principal models in terms of MSE recovery.
% All latent-space methods (M5--M8) are anchored at $z_1$, so their uncorrected output is $D(z_1)=y_2$ rather than $y_1$.
% Unlike the image-space methods M3 and M4, they must therefore recover the $y_1\!\to y_2$ degradation before improving upon the one-shot baseline.
For M5--M8, zero correction yields $D(z_1)=y_2$, so they must
recover the additional self-reconstruction degradation before
surpassing $y_1$.
At $N_{\mathrm{tr}}=2$, M6 outperforms the trained linear corrector
M5-T on all six datasets, supporting the benefit of nonlinear SRD-conditioned correction.
The $x$-free M6-xFree improves reconstruction on all six datasets, achieving $40.7\%$ average MSE recovery using only quantities derived from self-reconstruction.
To complement this normalized evaluation, Appendix
Table~\ref{tab:PSNR_gain} reports absolute reconstruction performance in PSNR.
M6-xFree improves PSNR by an average of $1.74$\,dB over the one-shot AE baseline,
compared with $2.01$\,dB for the supervised M6 ($N_{\mathrm{tr}}=2$).
The baseline AE requires $0.257$\,ms per ImageNet image.
M5 and M6-xFree ($N_{\mathrm{tr}}=2$) require approximately $2.9\times$ and $6.3\times$ the
baseline runtime, respectively, whereas computing the empirical decoder
reference $D(z^\ast)$ requires approximately $338$\,ms ($\sim1300\times$).
This reference is not required during SRD-RR inference.

Reconstruction examples and error maps are shown in Appendix \ref{app:image_results1}.
Across five random seeds and multiple AE architectures, SRD-RR consistently improves reconstruction across all settings.
Although the exact gains vary, the overall trend is preserved across initializations and changes in AE depth and channel capacity (Figure~\ref{fig:robustness} in Appendix).

% 5.6
\vspace{-5pt}
\subsection{Extension to a Pretrained Representation Autoencoder (RAE)}
\begin{table}[t]
\centering
\caption{RAE results for trajectory informativeness and pixel--perceptual reconstruction quality.}
\label{tab:RAE_results}
\setlength{\tabcolsep}{5pt}
\small
\begin{tabular}{lcccc}
\hline
Model & $N_{\rm tr}$ & PSNR $\uparrow$ & LPIPS $\downarrow$ & rFID-5k $\downarrow$ \\
\hline
M1
& -- & 18.87  & \textbf{0.156} & \textbf{3.91} \\

M4
& 1 & 17.70  & 0.201 & 5.72 \\
\hline
M6-LO
& 0 & 18.65  & 0.470 & 25.52 \\

M6
& 1 & \textbf{18.99} & 0.445 & 21.76 \\
\hline
M6-xFree
& 1 & 18.20 & 0.208 & 5.72 \\
\hline
\end{tabular}
\end{table}

% \begin{table}[t]
% \centering
% \caption{RAE results for trajectory informativeness and pixel--perceptual reconstruction quality.}
% \label{tab:RAE_results}
% \setlength{\tabcolsep}{5pt}
% \small
% \begin{tabular}{lccccc}
% \hline
% Model & $N_{\rm tr}$ & PSNR $\uparrow$ & $\Delta$PSNR & LPIPS $\downarrow$ & rFID-5k $\downarrow$ \\
% \hline
% M1
% & -- & 18.87 & 0.00 & \textbf{0.156} & \textbf{3.91} \\

% M4
% & 1 & 17.70 & $-1.17$ & 0.201 & 5.72 \\

% M6-LO
% & 0 & 18.65 & $-0.22$ & 0.470 & 25.52 \\

% M6
% & 1 & \textbf{18.99} & \textbf{+0.12} & 0.445 & 21.76 \\

% M6-xFree
% & 1 & 18.20 & $-0.67$ & 0.208 & 5.72 \\
% \hline
% \end{tabular}
% \end{table}
\vspace{-5pt}
Table~\ref{tab:RAE_results} summarizes the RAE results.
The key comparison for SRD informativeness is M6 versus M6-LO, which uses the same predictor backbone and pixel-MSE objective but removes the trajectory input. 
%
% Adding one SRD transition improves all five evaluated metrics 
% (Appendix Table~\ref{tab:rae_full_results}), including PSNR from 18.65 to 18.99 dB, LPIPS from 0.470 to 0.445, and rFID-5k from 25.52 to 21.76.
Adding one SRD transition improves all five evaluated metrics,
including PSNR by $0.341\,\mathrm{dB}$ and LPIPS by $0.0255$,
with paired 95\% confidence intervals excluding zero
(Appendix \ref{app:RAE_refinement}).
Thus, despite the substantially different latent geometry, SRD again provides useful information beyond $z_1$ alone.

The comparison with M1 reveals a separate issue.
M6 slightly improves pixel fidelity over M1 (PSNR: 18.87$\rightarrow$18.99 dB), but substantially degrades perceptual metrics (LPIPS: 0.156$\rightarrow$0.445; rFID-5k: 3.91$\rightarrow$21.76).
Importantly, the mismatch also occurs without trajectory conditioning;
adding SRD improves both perceptual metrics over M6-LO.
M6-xFree produces a more conservative refinement, improving PSNR over its pseudo-target baseline M4 (17.70$\rightarrow$18.20 dB) while remaining much closer to M1 in perceptual quality.
These results show that SRD informativeness persists in the pretrained RAE, while pixel-level reconstruction improvement and perceptual preservation are distinct objectives.

% Table~\ref{tab:RAE_results} summarizes the results on the DINOv2-based RAE.
% Repeated self-reconstruction again progressively degrades fidelity: PSNR decreases from 18.87~dB for $y_1$ to 14.97~dB for $y_5$, while rFID-5k increases from 3.91 to 33.44. 
% The latent SRD also exhibits substantially different and more strongly curved geometry than that of the convolutional AEs (Appendix~\ref{app:RAE_SRD}).
% %
% Trajectory conditioning nevertheless remains informative: M6
% ($N_{\rm tr}=1$) improves PSNR by 0.34~dB over M6-LO and by 0.12~dB over M1. 
% This pixel gain, however, is accompanied by substantial perceptual
% degradation (LPIPS: 0.156$\rightarrow$0.445; rFID-5k: 3.91$\rightarrow$21.76).
% In contrast, M6-xFree attains lower PSNR than M6
% (18.20 vs.\ 18.99~dB) but substantially better LPIPS (0.208) and rFID-5k (5.72), both much closer to M1. It also improves PSNR by 0.50~dB over M4 while largely preserving its perceptual quality. These results show that SRD remains informative in a substantially different pretrained AE, while the refinement target strongly affects the balance between pixel fidelity and perceptual preservation.
%
\section{Discussion}
    %\section{Discussion}
\label{sec:discussion}

%6.1
\vspace{-5pt}
\subsection{SRD as a Sample-Specific Reconstruction Signal}
\vspace{-5pt}
Our central finding is that repeated self-reconstruction itself does not improve reconstruction; nevertheless, the resulting SRD contains information useful for correcting the initial reconstruction.
M6 substantially outperforms the trajectory-free M6-LO, whereas assigning another sample's trajectory causes severe degradation.
Thus, SRD is useful not because it captures a generic drift of the AE, but because its dynamics are tied to the individual sample.

In the convolutional AEs, successive SRD steps are locally coherent but progressively deviate from the initial direction, while remaining partially aligned with the empirical reconstruction-improving direction.
This behavior is consistent with the effectiveness of simple reverse extrapolation at early steps, its limited benefit under repeated application, and the larger gains obtained by nonlinear SRD-RR.
It also agrees with the observation that most useful trajectory information is captured by the first few transitions.

%6.2
\vspace{-5pt}
\subsection{\texorpdfstring{From Image-Space Evidence to $x$-Free Refinement}{From Image-Space Evidence to x-Free Refinement}}
\vspace{-5pt}
The large image-space gains of M3 (+6.81\,dB) and M4 (+3.13\,dB) show that image-space SRD contains substantial reconstruction-relevant information.
However, these gains are largely lost after re-encoding because the existing encoder does not reliably preserve the corrected reconstruction.
This motivates predicting a latent correction directly and evaluating its output through the fixed decoder.
% However, most of these gains disappear after re-encoding because the image-space corrections are not constrained to remain compatible with the fixed decoder.
%This motivates exploiting SRD directly in latent space.
%
M6 performs such decoder-compatible refinement, achieving a 2.01\,dB average PSNR gain, but uses the original image $x$ as its refinement target.
M6-xFree instead uses the SRD-derived pseudo-target
$\tilde{x}_{\mathrm{lin}}$.
Once the initial self-reconstruction quantities have been generated, its refinement training and inference require no direct access to the original image $x$.
M6-xFree achieves a 1.74\,dB average PSNR gain and 40.7\% average MSE recovery across six datasets, without per-sample latent optimization.
These results show that substantial reconstruction improvement can be obtained using only information generated by the frozen AE's own self-reconstruction process.
The auxiliary-guide variants M7 and M8 provide only marginal or dataset-dependent additional gains over M6 and M6-xFree, respectively, suggesting that the short SRD already captures much of the useful information for refinement.
Detailed comparisons are provided in Appendix~\ref{app:aux_models}.

%6.3
\vspace{-5pt}
\subsection{Refinement Objectives and the Pretrained RAE}
\vspace{-5pt}
% The RAE experiment shows that the usefulness of SRD is not restricted to the latent geometry observed in the convolutional AEs.
% Despite exhibiting qualitatively different SRD geometry, the pretrained DINOv2-based RAE again undergoes progressive degradation under repeated self-reconstruction, and trajectory conditioning improves pixel-space reconstruction relative to a trajectory-free predictor.
% This suggests that the usefulness of SRD does not require a particular
% trajectory geometry.
% %
% At the same time, the RAE results expose an important dependence on the refinement objective.
% Although M6 improves PSNR, the perceptual image quality metrics LPIPS and rFID deteriorate significantly.
% In contrast, M6-xFree attains lower PSNR than M6 but substantially better LPIPS and rFID, although it does not surpass M1 on average.
% Thus, an SRD signal can remain informative even when optimizing it for one reconstruction criterion adversely affects another.
%
The RAE extension separates two questions that largely coincide in the conventional AEs: whether SRD is informative and whether the chosen refinement objective produces a desirable reconstruction.
The matched M6/M6-LO comparison answers the first: trajectory conditioning improves reconstruction prediction despite the qualitatively different SRD geometry. 
The comparison with M1 exposes the second: optimizing pixel-space MSE can improve pixel fidelity while substantially degrading perceptual quality.
Thus, SRD can remain a useful predictive signal even when the objective
used to exploit it is not aligned with perceptual reconstruction quality.

A related issue appears in the conventional AE experiments.
Directly regressing toward the empirically optimized latent $z^\ast$ (M6-Lobj) is substantially less stable than optimizing the decoded reconstruction as in M6.
This suggests that Euclidean proximity in latent space is not necessarily aligned with reconstruction quality, since decoder sensitivity can vary across latent directions.
This interpretation is consistent with work on decoder-induced latent geometry, in which Euclidean latent distances need not reflect distances between decoded outputs \citep{arvanitidis2017latent}.
Thus, $z^\ast$ serves as a useful evaluation reference, but is not necessarily an effective Euclidean regression target.
More generally, reconstruction fidelity alone may not determine whether a refined latent representation remains suitable for subsequent processing.
For example, StyleGAN inversion exhibits a distortion--editability trade-off, where improved reconstruction does not necessarily yield a latent representation suitable for semantic manipulation \citep{tov2021designing}.
Our RAE results similarly indicate that, for pretrained representation models, decoder compatibility alone may be insufficient: the refinement objective should also reflect the perceptual or downstream properties that the latent representation is expected to preserve.

%6.4
\vspace{-5pt}
\subsection{Limitations and Future Work}
\vspace{-5pt}
Our main evaluation uses convolutional deterministic AEs across six datasets, while the pretrained-model extension is limited to one DINOv2-based RAE on ImageNet100.
Further evaluation across broader AE families is therefore needed.
The relation between SRD and decoder geometry also remains incompletely understood.
Future work includes analyzing SRD through local decoder geometry, developing objectives that better balance pixel fidelity and perceptual quality, improving $x$-free refinement, and examining compatibility with downstream latent-space operations.
More broadly, we aim to understand whether naturally occurring transient dynamics in pretrained models can serve as intrinsic, sample-specific signals for representation refinement.
\section{Conclusion}
%\section{Conclusion}
\vspace{-3pt}
We investigated whether a trained AE contains exploitable information for improving its own initial reconstruction.
Self-reconstruction can degrade image fidelity while exposing dynamics useful for refinement.
SRD-RR exploits these dynamics through a learned latent correction with a frozen AE and no per-sample test-time optimization.
Across six datasets with convolutional AEs, nonlinear SRD-RR outperforms linear latent extrapolation, including in an $x$-free variant trained on an SRD-derived pseudo-target. Trajectory ablations support the practical value of correctly paired SRD features.
The RAE extension further supports SRD informativeness under different latent dynamics, while showing that pixel fidelity and perceptual quality are distinct. 
Together, these findings establish self-reconstruction trajectories as useful signals for refinement with frozen autoencoders.

% The pretrained RAE extension reveals a limitation: improved pixel fidelity need not preserve perceptual quality. 
% Future refinement objectives should therefore reflect the properties required of the reconstructed image and latent representation.

%
% \subsubsection*{Author Contributions}
% If you'd like to, you may include  a section for author contributions as is done
% in many journals. This is optional and at the discretion of the authors.

% \subsubsection*{Acknowledgments}
% Use unnumbered third level headings for the acknowledgments. All
% acknowledgments, including those to funding agencies, go at the end of the paper.

\section*{Reproducibility Statement}
To support reproducibility, we report the dataset splits and evaluation protocol in Section~4.1 and Table~2. 
Appendix~C provides detailed preprocessing, model architectures, training procedures, hyperparameters, latent normalization, and the empirical decoder-optimized reference.
Appendix~D provides the pretrained RAE configuration and its evaluation protocol, including perceptual metrics and confidence-interval computation.

\section*{Use of Generative AI}
Generative AI tools were used to assist with language editing, literature
survey, manuscript organization, code development and debugging, and discussion
of experimental analyses.
All AI-assisted code and analyses were reviewed and verified by the authors.

The authors made all final decisions regarding the research methodology,
experimental design, interpretation of results, and conclusions, and take full
responsibility for the content of this work.

%\bibliography{iclr2027_conference}
\bibliography{main}
\bibliographystyle{iclr2027_conference}

\newpage

\appendix

\setcounter{figure}{0}
\renewcommand{\thefigure}{A\arabic{figure}}
\setcounter{table}{0}
\renewcommand{\thetable}{A\arabic{table}}

\section{Additional Results and Discussion}
%\section{Additional Results}
\label{app:results}
\FloatBarrier

\begin{table*}[!b]
\centering
\caption{
PSNR improvements over the baseline reconstruction and corresponding MSE recovery ratios. M1 reports the absolute PSNR, whereas all other rows report the PSNR improvement ($\Delta$PSNR) relative to M1.
%RAE results for trajectory informativeness and pixel--perceptual reconstruction quality.
}
\label{tab:PSNR_gain}

\setlength{\tabcolsep}{2.5pt}
\small
\begin{tabular}{lcrrrrrrrrc}
\hline
\multirow{2}{*}{Method}
& \multirow{2}{*}{$\mathbf{N_{\mathrm{tr}}}$}
& \multirow{2}{*}{CIFAR-10}
& \multirow{2}{*}{SVHN}
& \multirow{2}{*}{FMNIST}
& \multirow{2}{*}{MNIST}
& \multirow{2}{*}{ImageNet}
& \multirow{2}{*}{CelebA}
& \multicolumn{1}{c}{macro}
& \multicolumn{1}{c}{macro}
& \multirow{2}{*}{$x$-free} \\
& & & & & & & 
& $\Delta$PSNR
& {\tiny MSE-recov. (\%)}
& \\
\hline

M1
& -- 
& 30.42 & 36.91 & 32.37 & 35.33 & 35.03 & 38.33
& 0.00 & 0.00 & -- \\
\hline

M3 $^\dagger$
& 1 
& 3.80 & 7.94 & 8.30 & 12.75 & 2.00 & 6.09
& 6.81 & 92.38 &\\
%91.18 &  \\

M4 $^\dagger$
& 1 
& 1.36 & 3.59 & 3.98 & 4.28 & 2.67 & 2.90
& 3.13 & 62.19 & \checkmark  \\
%& 58.86 & \checkmark  \\
\hline

M5
& 1
& 0.57 & 1.56 & 0.84 & 0.61 & 1.19 & 0.96
& 0.95 & 24.62 & \checkmark  \\
%& 18.97 & \checkmark  \\
\hline

M6-LO
& 0
& -0.24 & 0.94 & -1.30 & -2.57 & 0.28 & -0.77
& -0.61 & -24.32 & \\
%& -29.55 & \\

M6
& 1
& 0.95 & 2.80 & 1.74 & 0.86 & 1.66 & 1.60
& 1.60 & 38.55 &  \\
%& 34.18 &  \\

M6
& 2
& 1.05 & 3.08 & 2.18 & 1.88 & 1.87 & 2.00
& 2.01 & 45.32 &  \\
%& 2.01 & 41.09 &  \\

M6
& 4
& 1.06 & 3.06 & 2.19 & 1.96 & 1.93 & 2.07
& 2.04 & 44.93 &  \\
%& 2.04 & 41.40 &  \\ ******************

M6-Lobj
& 2
& 0.85 & 2.44 & -5.46 & -22.45 & 1.71 & -2.18
& -4.18 & -3257.76 &  \\
%& -4.18 & -2966.13 &  \\

\textbf{M6-xFree}
& \textbf{2}
& \textbf{0.87} & \textbf{2.53} & \textbf{1.89} & \textbf{1.66}
& \textbf{1.84} & \textbf{1.66}
& \textbf{1.74} 
%& \textbf{35.94} & \checkmark  \\
& \textbf{40.65} & \checkmark  \\

M7
& 2
& 1.08 & 3.25 & 2.73 & 2.58 & 1.08 & 1.88
& 2.10 & 45.16 &  \\
%& 2.10 & 41.53 &  \\

M8
& 2
& 0.91 & 2.54 & 1.93 & 1.73 & 1.83 & 1.64
%& 1.76 & 36.53 & \checkmark  \\
& 1.76 & 41.17 & \checkmark  \\
\hline

$D(z^{*})$
& --
& 3.01 & 7.11 & 13.17 & 16.58 & 5.22 & 6.09
& 8.53 & 100.00 & -- \\
\hline
\end{tabular}

{\scriptsize
\raggedright
$^\dagger$: M3 and M4 improve image quality without enforcing decoder compatibility.
\par}
\end{table*}

Table~\ref{tab:PSNR_gain} summarizes the main reconstruction results in PSNR and the corresponding improvements over the baseline.

% A.1
\subsection{Linear Image-Space and Latent-Space Correctors}

\begin{figure*}[t]
    \centering

    \begin{subfigure}[t]{0.48\textwidth}
        \centering
        \includegraphics[width=\linewidth]{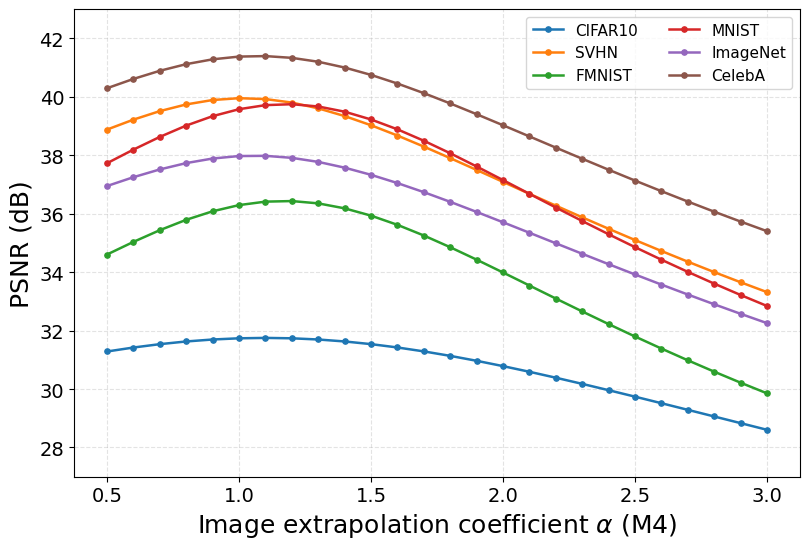}
        \caption{$\alpha$ sweep}
        \label{fig:alpha_sweep}
    \end{subfigure}
    \hfill
    \begin{subfigure}[t]{0.48\textwidth}
        \centering
        \includegraphics[width=\linewidth]{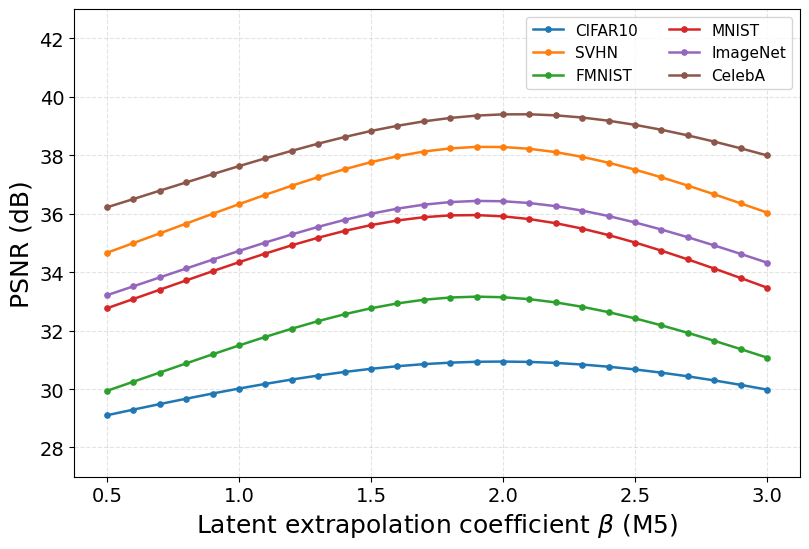}
        \caption{$\beta$ sweep}
        \label{fig:beta_sweep}
    \end{subfigure}

    \caption{Validation-set performance of image-space and latent-space linear
    extrapolation under sweeps of $\alpha$ and $\beta$.}
    \label{fig:alpha_beta_sweep}
\end{figure*}

Figure~\ref{fig:alpha_beta_sweep} shows validation-set PSNR sweeps for the image-space (M4) and latent-space (M5) linear extrapolation methods.
For M4, we fix $\alpha=1.0$ based on the direct linear extrapolation formulation; the validation sweep confirms that this choice is near-optimal across datasets.
For M5, the validation sweeps consistently favor $\beta\approx2.0$ across all six datasets, and we therefore use $\beta=2.0$ for final test evaluation.
Notably, the first coefficient independently learned by M5-T is also close to 2.0 across datasets.

Because M5 is training-free, computationally efficient, decoder-compatible, and $x$-free, we additionally examine repeated application of M5.

\begin{figure}[t]
    \centering
    \includegraphics[width=0.6\columnwidth]{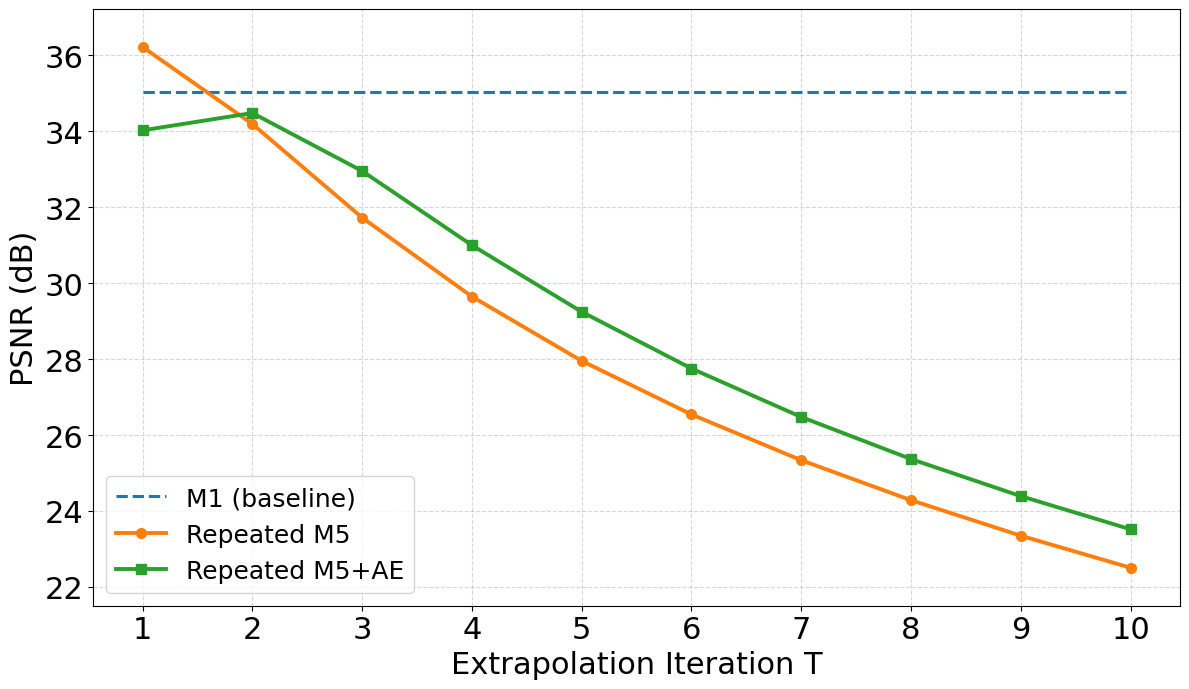}
    \caption{Performance of Repeated M5 and Repeated M5+AE on ImageNet.}
    \label{fig:iterativeM5}
\end{figure}

Figure~\ref{fig:iterativeM5} shows the performance of Iterative M5 and Iterative M5-AE, in which an AE step is applied after Iterative M5.
Repeated M5 correction progressively degrades reconstruction quality.
This behavior is consistent with the changing direction of the latent SRD and the limited validity of repeated linear extrapolation, motivating nonlinear SRD-RR methods such as M6 and M6-xFree.

% A.2
\subsection{Latent SRD Trajectories}

\begin{figure}[t]
    \centering
    % ---- 1st row ----
    \begin{subfigure}[t]{0.32\linewidth}
        \centering
        \includegraphics[width=\linewidth]{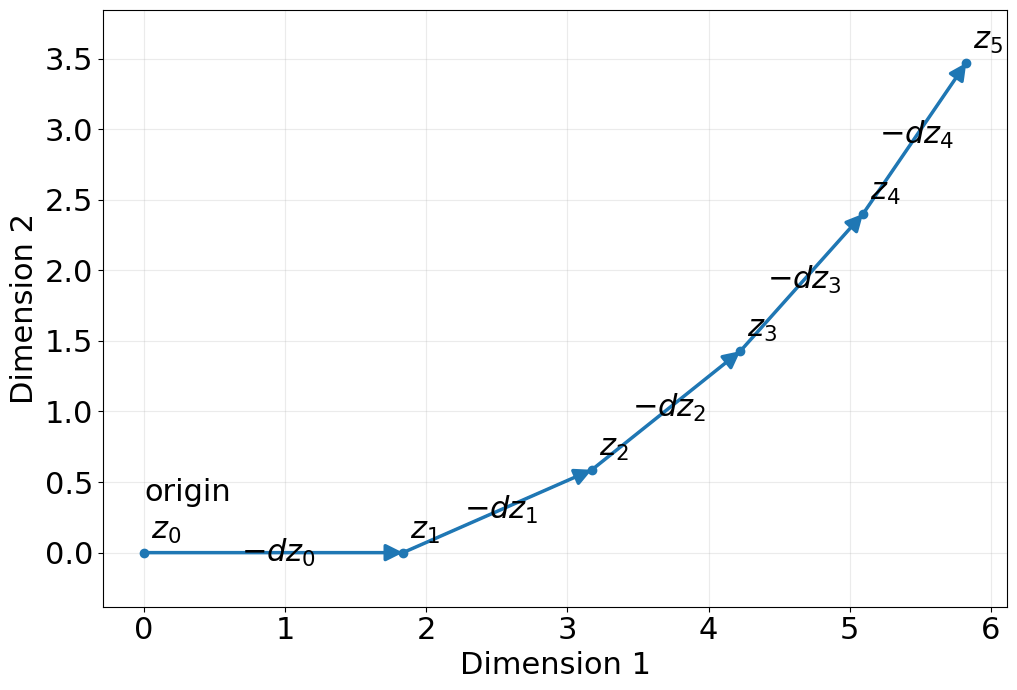}
        \caption{CIFAR-10}
        \label{fig:traj_cifar10}
    \end{subfigure}
    \hfill
    \begin{subfigure}[t]{0.32\linewidth}
        \centering
        \includegraphics[width=\linewidth]{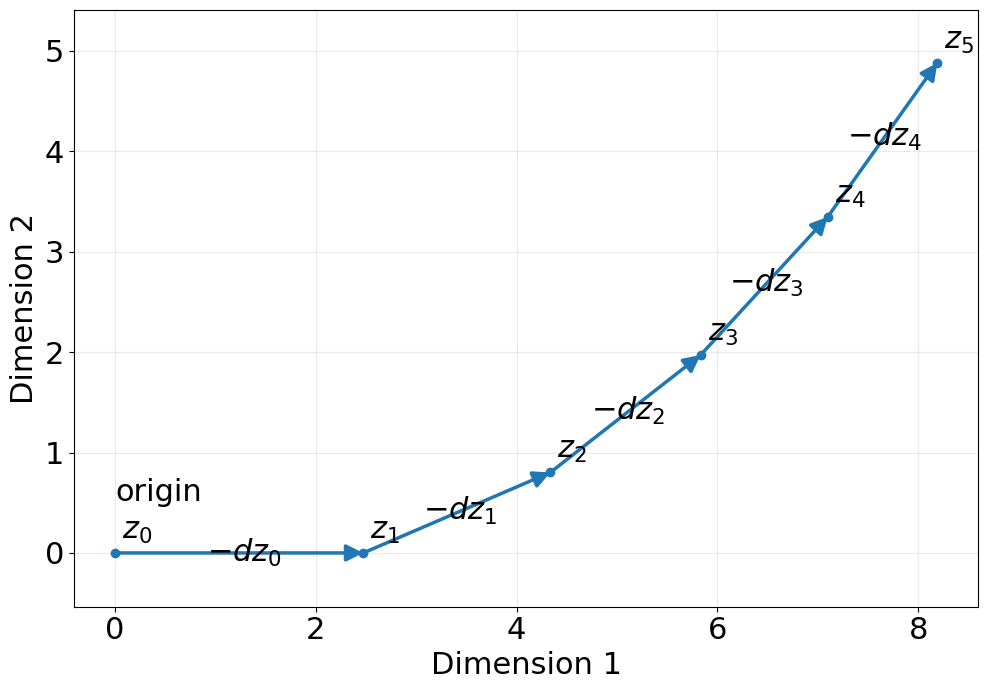}
        \caption{SVHN}
        \label{fig:traj_svhn}
    \end{subfigure}
    \hfill
    \begin{subfigure}[t]{0.32\linewidth}
        \centering
        \includegraphics[width=\linewidth]{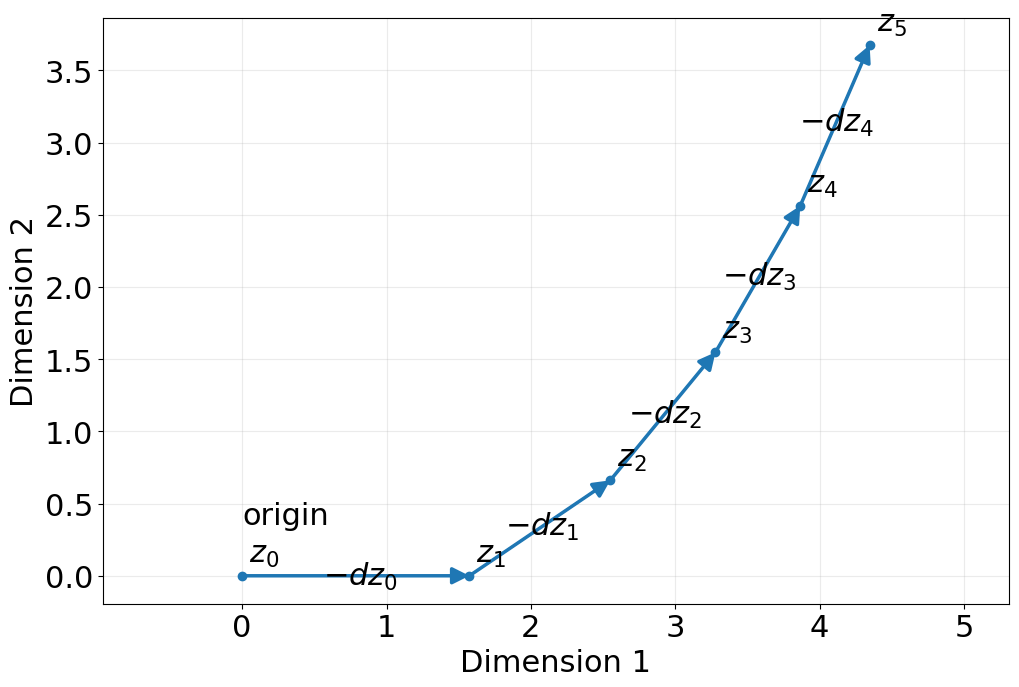}
        \caption{FashionMNIST}
        \label{fig:traj_fmnist}
    \end{subfigure}

    \vspace{0.8em}

    % ---- 2nd row ----
    \begin{subfigure}[t]{0.32\linewidth}
        \centering
        \includegraphics[width=\linewidth]{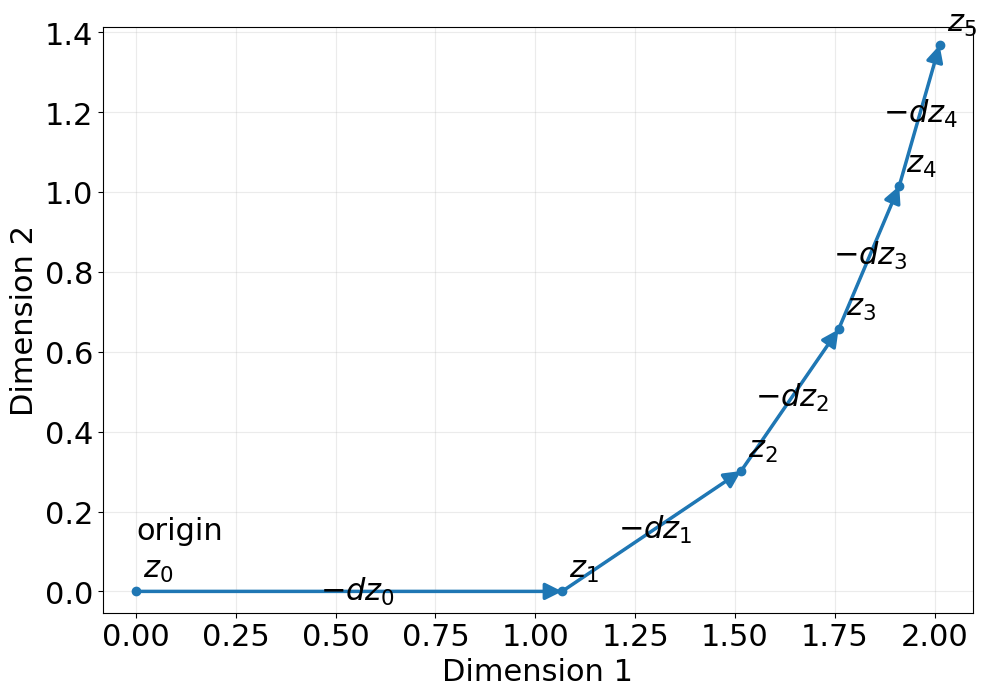}
        \caption{MNIST}
        \label{fig:traj_mnist}
    \end{subfigure}
    \hfill
    \begin{subfigure}[t]{0.32\linewidth}
        \centering
        \includegraphics[width=\linewidth]{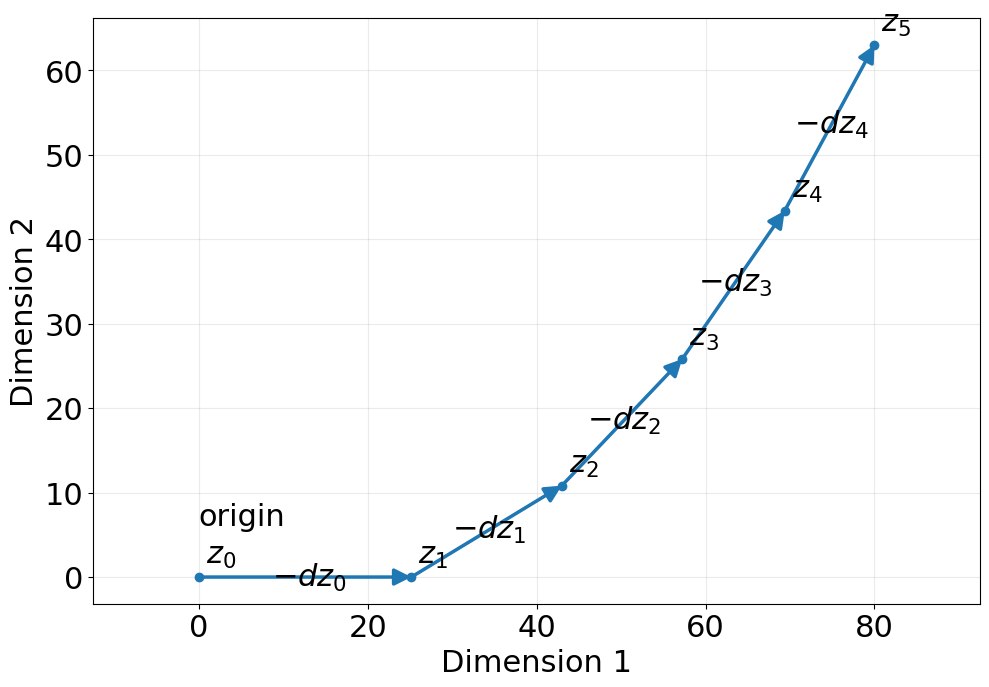}
        \caption{ImageNet}
        \label{fig:traj_imagenet}
    \end{subfigure}
    \hfill
    \begin{subfigure}[t]{0.32\linewidth}
        \centering
        \includegraphics[width=\linewidth]{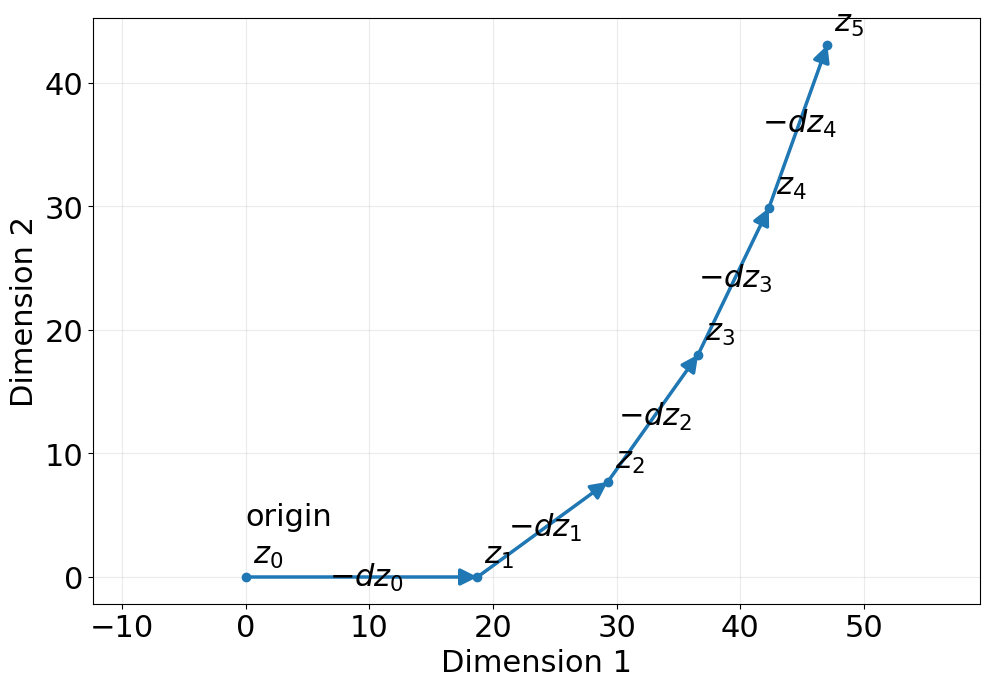}
        \caption{CelebA}
        \label{fig:traj_celeba}
    \end{subfigure}

    \caption{Changes in the latent trajectories across the six datasets.}
    \label{fig:trajectory_all}
\end{figure}

Figure~\ref{fig:trajectory_all} visualizes the latent SRD trajectories in two dimensions using the observed magnitudes and cosine similarities of the $dz_k$ vectors.
Vector directions are fitted by least squares to reproduce the observed cosine similarities while preserving their magnitudes.
The fitting error is small across all datasets, with a mean cosine-similarity error of (0.0162 $\pm$ 0.0164).

Thus, the two-dimensional representations faithfully visualize the directional changes observed in the original high-dimensional trajectories.
These results are consistent with the limitations of global linear extrapolation and the degradation observed under repeated M5 correction.

% A.3
\subsection{Latent-Objective and Auxiliary-Guide Ablations}
\label{app:aux_models}

Directly regressing toward the empirically optimized latent $z^\ast$ (M6-Lobj) is substantially less stable than optimizing the decoded reconstruction as in M6.
This suggests that Euclidean proximity in latent space is not necessarily aligned with reconstruction quality, since decoder sensitivity can vary across latent directions.
Thus, $z^\ast$ is useful as an evaluation reference but is not necessarily an effective Euclidean regression target.

The auxiliary-guide experiments (M7 and M8) examine whether additional latent guides provide useful information beyond the SRD trajectory.
Under original-image supervision, M7 augments M6 with a latent guide derived from the M3 correction.
Its average PSNR gain is 2.10\,dB, compared with 2.01\,dB for M6, while its average MSE-recov is 45.2\%, essentially comparable to M6.
Its effect is also dataset-dependent, including degradation on ImageNet.
Thus, the additional image-derived guide is not uniformly beneficial.

In the $x$-free setting, M8 augments M6-xFree with
$\tilde{z}_{0,\mathrm{lin}}=E(\tilde{x}_{\mathrm{lin}})$ while using the same pseudo-target.
M8 achieves an average PSNR gain of 1.76\,dB and an average MSE-recov of 41.2\%, compared with 1.74\,dB and 40.7\% for M6-xFree.
The improvement is therefore small despite increasing the computational cost from $6.3\times$ to $6.9\times$ M1.
This matched comparison suggests that the additional guide provides only limited benefit beyond the SRD trajectory and its SRD-derived training target.

% A.4
\subsection{Robustness}
\label{sec:result_robustness}

\begin{figure*}[t]
    \centering

    \begin{subfigure}[t]{0.325\textwidth}
        \centering
        \includegraphics[width=\linewidth]{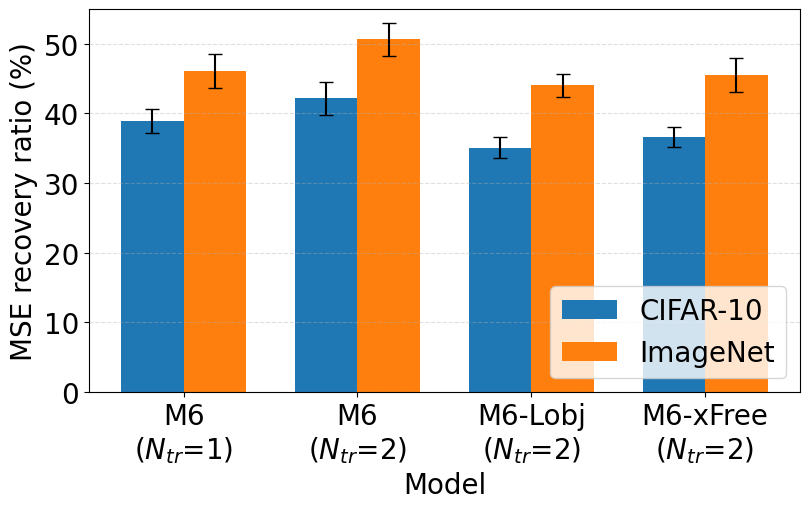}
        \caption{Five different seeds}
        \label{fig:five_seeds}
    \end{subfigure}
    \hfill
    \begin{subfigure}[t]{0.325\textwidth}
        \centering
        \includegraphics[width=\linewidth]{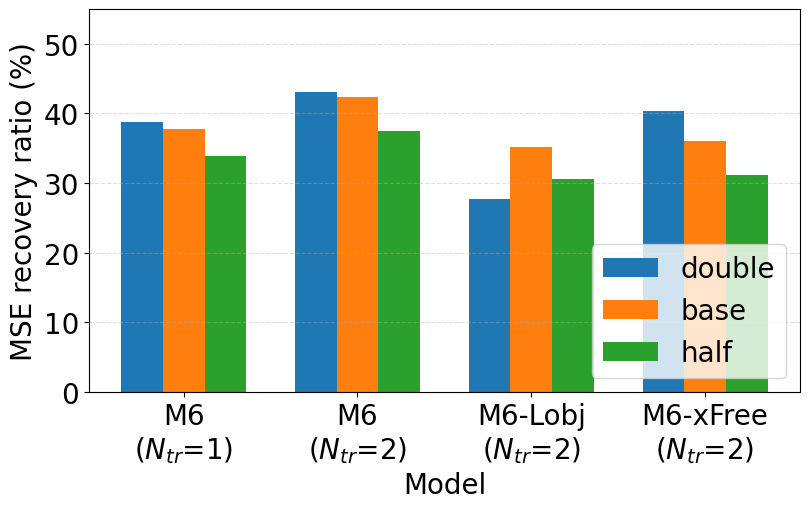}
        \caption{CIFAR-10}
        \label{fig:arch_cifar}
    \end{subfigure}
    \hfill
    \begin{subfigure}[t]{0.325\textwidth}
        \centering
        \includegraphics[width=\linewidth]{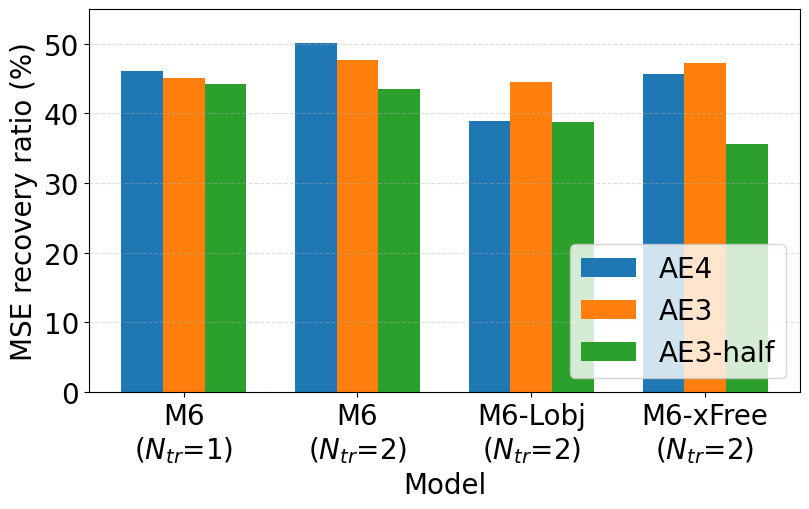}
        \caption{ImageNet}
        \label{fig:arch_imagenet}
    \end{subfigure}

    \caption{Performance consistency across (a) five random seeds and (b)(c) model architectures.}
    \label{fig:robustness}
\end{figure*}

Figure~\ref{fig:robustness} evaluates SRD-RR across five random seeds and multiple AE architectures on CIFAR-10 and ImageNet.
In (a), bars show means and error bars show standard deviations across five seeds.
For each seed, both the AE and SRD-RR predictor are retrained from scratch using a seed-specific training/validation split, while the test set is kept fixed.
SRD-RR consistently improves reconstruction across all settings.
Although the exact gains vary, the overall trend is preserved across random initializations and changes in AE depth and channel capacity.

\section{Qualitative Results}
%\section{Qualitative Results}
\label{app:image_results1}
%\FloatBarrier
\begin{figure*}[t]
    \centering

    % ===== Top row =====
    \begin{subfigure}[t]{0.23\textwidth}
        \centering
        \includegraphics[width=\linewidth]{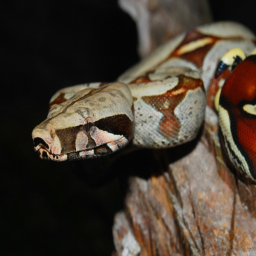}
        \caption{GT}
        \label{fig:gt}
    \end{subfigure}
    \hfill
    \begin{subfigure}[t]{0.23\textwidth}
        \centering
        \includegraphics[width=\linewidth]{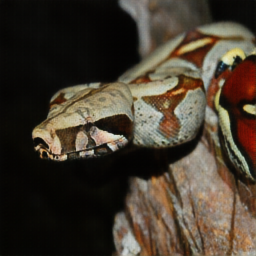}
        \caption{M1}
        \label{fig:Ex1_m1}
    \end{subfigure}
    \hfill
    \begin{subfigure}[t]{0.23\textwidth}
        \centering
        \includegraphics[width=\linewidth]{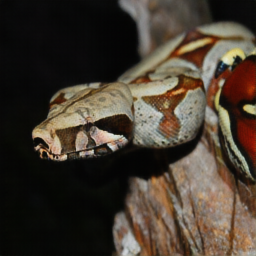}
        \caption{M6}
        \label{fig:Ex1_m6}
    \end{subfigure}
    \hfill
    \begin{subfigure}[t]{0.23\textwidth}
        \centering
        \includegraphics[width=\linewidth]{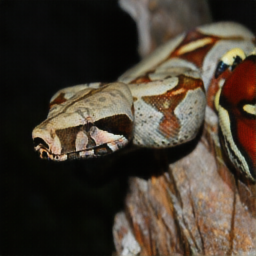}
        \caption{M6-xFree}
        \label{fig:Ex1_m6d}
    \end{subfigure}

    \vspace{1mm}

    % ===== Bottom row =====
    \begin{subfigure}[t]{0.23\textwidth}
        \centering
        \includegraphics[width=\linewidth]{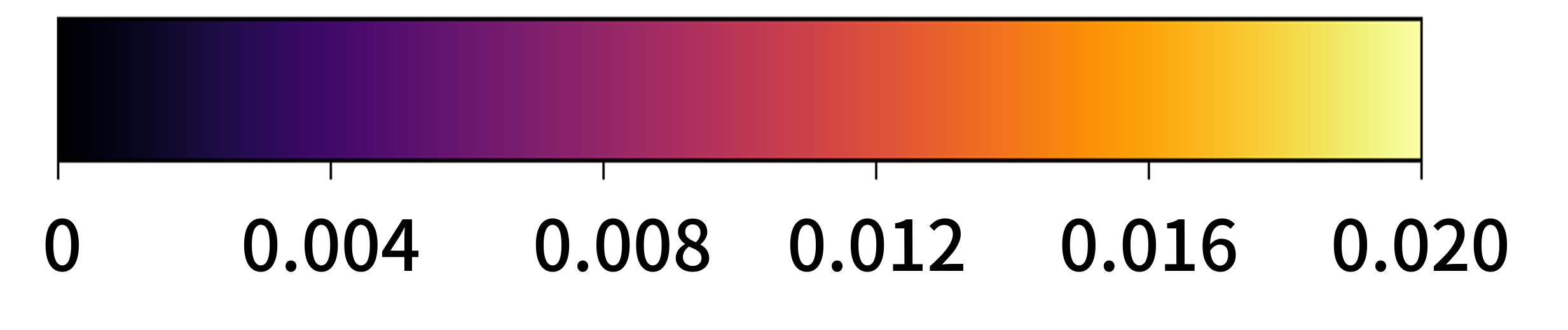}
        \caption{Error intensity}
        \label{fig:colorbar1}
    \end{subfigure}
    \hfill
    \begin{subfigure}[t]{0.23\textwidth}
        \centering
        \includegraphics[width=\linewidth]{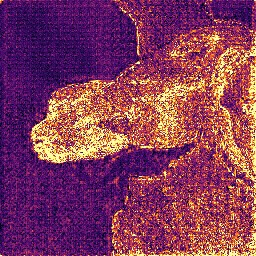}
        \caption{$|$M1 $-$ GT$|$}
        \label{fig:Ex1_diff_m1}
    \end{subfigure}
    \hfill
    \begin{subfigure}[t]{0.23\textwidth}
        \centering
        \includegraphics[width=\linewidth]{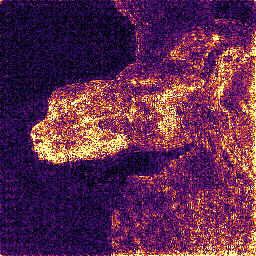}
        \caption{$|$M6 $-$ GT$|$}
        \label{fig:Ex1_diff_m6}
    \end{subfigure}
    \hfill
    \begin{subfigure}[t]{0.23\textwidth}
        \centering
        \includegraphics[width=\linewidth]{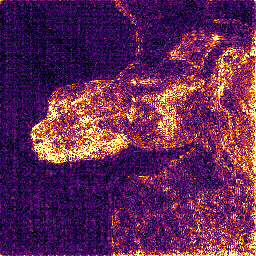}
        \caption{$|$M6-xFree $-$ GT$|$}
        \label{fig:Ex1_diff_m6d}
    \end{subfigure}

    \caption{
    Qualitative comparison of reconstruction results and corresponding error maps from ImageNet.
    %The top row shows the ground truth (GT) and reconstructed images obtained by
    M1 (37.76dB), M6 (39.71dB, MSE-recov=51.3\%), and M6-xFree (39.68dB, MSE-recov=50.5\%).
    %The bottom row shows the error-intensity color scale ($\times$20) and the absolute
    %difference maps between each reconstruction and the GT.
    }
    \label{fig:image_examples1}
\end{figure*}
\begin{figure*}[t]
    \centering

    % ===== Top row =====
    \begin{subfigure}[t]{0.23\textwidth}
        \centering
        \includegraphics[width=\linewidth]{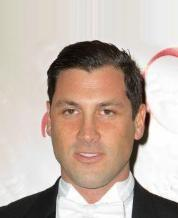}
        \caption{GT}
        \label{fig:Ex2_gt}
    \end{subfigure}
    \hfill
    \begin{subfigure}[t]{0.23\textwidth}
        \centering
        \includegraphics[width=\linewidth]{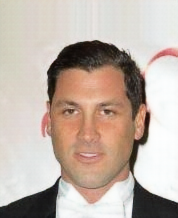}
        \caption{M1}
        \label{fig:Ex2_m1}
    \end{subfigure}
    \hfill
    \begin{subfigure}[t]{0.23\textwidth}
        \centering
        \includegraphics[width=\linewidth]{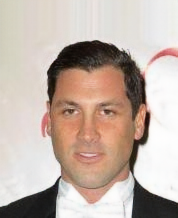}
        \caption{M6}
        \label{fig:Ex2_m6}
    \end{subfigure}
    \hfill
    \begin{subfigure}[t]{0.23\textwidth}
        \centering
        \includegraphics[width=\linewidth]{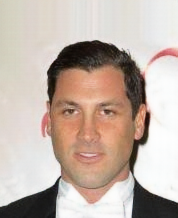}
        \caption{M6-xFree}
        \label{fig:Ex2_m6d}
    \end{subfigure}

    \vspace{1mm}

    % ===== Bottom row =====
    \begin{subfigure}[t]{0.23\textwidth}
        \centering
        \includegraphics[width=\linewidth]{images/ex/colorbar_x50.png}
        \caption{Error intensity}
        \label{fig:colorbar2}
    \end{subfigure}
    \hfill
    \begin{subfigure}[t]{0.23\textwidth}
        \centering
        \includegraphics[width=\linewidth]{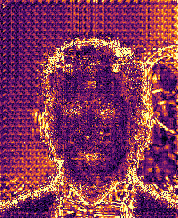}
        \caption{$|$M1 $-$ GT$|$}
        \label{fig:Ex2_diff_m1}
    \end{subfigure}
    \hfill
    \begin{subfigure}[t]{0.23\textwidth}
        \centering
        \includegraphics[width=\linewidth]{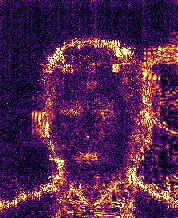}
        \caption{$|$M6 $-$ GT$|$}
        \label{fig:Ex2_diff_m6}
    \end{subfigure}
    \hfill
    \begin{subfigure}[t]{0.23\textwidth}
        \centering
        \includegraphics[width=\linewidth]{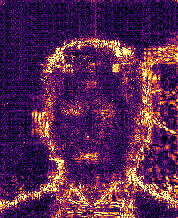}
        \caption{$|$M6-xFree $-$ GT$|$}
        \label{fig:Ex2_diff_m6d}
    \end{subfigure}

    \caption{
    Qualitative comparison of reconstruction results and corresponding error maps from CelebA.
    %The top row shows the ground truth (GT) and reconstructed images obtained by
    M1 (38.39dB), M6 (40.97dB, MSE-recov=55.6\%), and M6-xFree (40.50dB, MSE-recov=47.7\%).
    %The bottom row shows the error-intensity color scale ($\times$20) and the absolute
    %difference maps between each reconstruction and the GT.
    }
    \label{fig:image_examples2}
\end{figure*}
Figs.~\ref{fig:image_examples1} and ~\ref{fig:image_examples2} show reconstruction examples and the corresponding error maps, where the errors are magnified by a factor of 50, clipped to $[0,1]$, and colorized for visualization.

% show reconstruction examples and corresponding error maps, magnifying $\times$50 to [0, 1] and colorization for visualization purpose.

\section{Implementation Details for the main experiments}
%\section{Implementation Details}
\label{app:implementation}
%\FloatBarrier

%C.1
\subsection{Image Pre-processing}
\label{app:pre-processing}
CIFAR-10 and SVHN images were used at their native $32\times32$ resolution,
while MNIST and FashionMNIST were used at $28\times28$.
ImageNet100 images were resized to $256\times256$ using bilinear interpolation.
CelebA images were used at their native $218\times178$ resolution.
All images were converted to tensors with pixel values scaled to $[0,1]$; no additional normalization or data augmentation was applied.

Because the CelebA image dimensions are not divisible by the overall downsampling factor of the encoder, each $218\times178$ image was zero-padded on the bottom and right to $224\times184$ before being processed by the AE.
After decoding, the reconstructed image was cropped back to the original
$218\times178$ resolution before evaluation.

% C.2
\subsection{Autoencoder}
We use fully convolutional AEs with spatial latent representations.
Each encoder stage consists of a $3\times3$ convolution with stride 2 and
padding 1, followed by batch normalization \citep{ioffe2015batch} and
ReLU \citep{nair2010rectified}.
The decoder is symmetric, using transposed convolutions in reverse channel
order, with a sigmoid output layer.
The encoder channel dimensions are $[16,32,64]$ for CIFAR-10, SVHN, FMNIST,
and MNIST, and $[32,64,128]$ for ImageNet and CelebA.
The AE architecture is independent of the SRD length $K$.

For the architecture-consistency experiments, we additionally vary the AE
capacity while keeping the remaining protocol unchanged.
For CIFAR-10, we use encoder channels $[8,16,32]$ (half),
$[16,32,64]$ (standard), and $[32,64,128]$ (double).
For ImageNet, we use $[16,32,64]$ (half), $[32,64,128]$ (standard),
and a deeper four-stage architecture $[32,64,128,256]$ (AE-4).
Each decoder is symmetric to its corresponding encoder.

The AE is trained to minimize pixel-space MSE using Adam with a learning rate
of $10^{-3}$ and no weight decay.
We train for at most 50 epochs with early-stopping patience 5 and gradient clipping at norm 10.
The batch size is 64 for the four smaller datasets and 256 for ImageNet and CelebA.
Training uses FP16 automatic mixed precision (AMP) with FP32 master weights.
After training, all AE parameters and BatchNorm running statistics are kept fixed throughout SRD generation and all refinement experiments.
% C.3
\subsection{Image-Space Corrector}

M3 concatenates the one-shot reconstruction and the observed image-space SRD,
\[
    [y_1,dy_1,\ldots,dy_{K-1}],
\]
along the channel dimension.
It consists of an initial $3\times3$ convolution with 128 hidden channels,
three residual blocks \citep{he2016deep}, and a zero-initialized
$3\times3$ output convolution.
Each residual block contains two convolutions with batch normalization and a
residual connection.
The network predicts a residual $\Delta x$, giving
\[
    \tilde{x}=y_1+\Delta x.
\]

M3 is trained against the residual target $x-y_1$ using MSE.
We use Adam \citep{kingma2014adam} with learning rate $10^{-3}$,
no weight decay, at most 50 epochs, early-stopping patience 6, and gradient
clipping at norm 1.
The batch size is 64 for the smaller datasets and 256 for ImageNet and CelebA.
Training uses FP16 AMP.

% C.4
\subsection{Trained Linear Trajectory Corrector}
M5-T learns the coefficients $\{\beta_k^\ast\}$ by minimizing the pixel-space MSE between $D(z_1+\sum_k \beta_k^\ast dz_k)$ and the ground-truth image.
% We use Adam with a learning rate of 0.03, batch size 64, and train for at most 20 epochs with early-stopping patience 5 and a minimum improvement tolerance of $10^{-7}$.
We train M5-T using Adam with a learning rate of 0.03 and a batch size of 64 for at most 20 epochs, with early-stopping patience 5 and a minimum improvement tolerance of $10^{-7}$.

% C.5
\subsection{Latent-Space SRD-RR Predictor and Hyperparameters}

The nonlinear SRD-RR predictor takes the channel-wise concatenation
\[
    [z_1,dz_1,\ldots,dz_{K-1}]
\]
as input and predicts a latent correction $\Delta z$.
For CIFAR-10, SVHN, FMNIST, and MNIST, the predictor uses 64 hidden channels
and four residual blocks; for ImageNet and CelebA, it uses 128 hidden channels and five residual blocks.
The input and output projections use $3\times3$ convolutions, and each residual block consists of two $3\times3$ convolutions with batch normalization.
The final correction layer is initialized to zero.

The corrected latent representation and reconstruction are
\[
    \hat{z}=z_1+\gamma\Delta z,
    \qquad
    \hat{x}=D(\hat{z}),
\]
where $\gamma$ is a learnable scalar initialized to $0.5$.
All nonlinear SRD-RR variants use the same predictor backbone and training hyperparameters unless otherwise stated.
M6-LO uses the same predictor architecture and training settings as M6, except that its input projection receives only $z_1$ and therefore has fewer input channels.

For supervised SRD-RR, the predictor minimizes pixel-space MSE between
$D(\hat{z})$ and the original image $x$.
The $x$-free variant instead uses the SRD-derived pseudo-target
\[
    \tilde{x}_{\mathrm{lin}}
    = y_1+(y_1-y_2)
\]
and minimizes the same pixel-space MSE with
$\tilde{x}_{\mathrm{lin}}$ as the target.
Once the self-reconstruction quantities have been generated, M6-xFree requires no direct access to the original image $x$ during refinement training or inference.
For the $x$-free variants (M6-xFree and M8), early stopping uses only
the pseudo-target validation loss.
We optimize the SRD-RR predictors using AdamW
\citep{loshchilov2017decoupled} with learning rate $5\times10^{-4}$ and weight decay $10^{-3}$.
Training is performed for at most 150 epochs with early-stopping patience 10 and gradient clipping at norm 1.
The batch size is 64 for the smaller datasets and 256 for ImageNet and CelebA.
Pixel-objective predictors are trained using FP16 AMP, whereas the latent-objective M6-Lobj is optimized in FP32 for numerical stability.

% Table A2
\begin{table}[t]
\centering
\caption{Training hyperparameters for the learned components.
Batch sizes are shown as small/large datasets.}
\label{tab:implementation_hyperparameters}
\small
\setlength{\tabcolsep}{4pt}
\begin{tabular}{lccc}
\toprule
 & AE & M3 & SRD-RR \\
\midrule
Optimizer              & Adam  & Adam  & AdamW \\
Learning rate           & $10^{-3}$ & $10^{-3}$ & $5\times10^{-4}$ \\
Weight decay            & 0 & 0 & $10^{-3}$ \\
Batch size              & 64 / 256 & 64 / 256 & 64 / 256 \\
Max.\ epochs            & 50 & 50 & 150 \\
Early-stop patience     & 5 & 6 & 10 \\
Gradient clipping       & 10 & 1 & 1 \\
Objective               & pixel MSE & residual MSE & pixel MSE \\
Precision               & FP16 AMP & FP16 AMP & FP16 AMP$^\dagger$ \\
\bottomrule
\multicolumn{4}{l}{\footnotesize M6-Lobj uses a latent-space MSE objective and FP32 optimization.}
\end{tabular}
\end{table}
Detailed hyperparameters are summarized in
Table~\ref{tab:implementation_hyperparameters}.

% C.6
\subsection{Latent Normalization}

The latent representation $z_1$ and trajectory differences $dz_k$ are standardized per dimension using statistics estimated from the training set.
Statistics for $z_1$ are estimated separately, while the $dz_k$ share common per-dimension normalization statistics across trajectory transitions.
The predicted correction is rescaled using the trajectory-difference statistics before being added to $z_1$.
The same normalization is used during training and inference.

For trajectory storage, $z_1$ is stored in FP16, whereas $dz_k$ is stored in FP32.
Other latent states are reconstructed from these quantities when required.

% C.6
\subsection{Empirical Decoder-Optimized Reference}
\label{app:decoder_reference}

For each sample, the empirical decoder reference is obtained by solving
\[
    z^\ast =
    \arg\min_z \|D(z)-x\|_2^2
\]
with the decoder fixed and initialization $z_0=E(x)$.
We use Adam with learning rate $0.01$ for at most 500 optimization steps, gradient clipping at 14, and early stopping.
No latent-anchor regularization is used.
Optimization is performed in FP32 and batched only for computational efficiency.

In all evaluated samples, the optimized reference $D(z^\ast)$ achieved lower reconstruction error than the one-shot baseline $y_1$; no sample with a non-positive MSE-recov denominator was observed.
This optimization is used to construct the empirical evaluation reference and, for M6-Lobj, the latent regression target $z^\ast$ on the training data.
For test samples, $z^\ast$ is used only for evaluation and analysis.
No per-sample latent optimization is required by SRD-RR at inference time.

\section{Extension to a Pretrained Representation Autoencoder}
%\section{Appendix D: Extension to a Pretrained Representation Autoencoder}
\label{app:RAE}

% D.1
\subsection{Experimental Setup}
\label{app:RAE_setup}
To examine whether SRD extends beyond the convolutional deterministic AEs used in the main experiments, we additionally evaluate a pretrained DINOv2-based Representation Autoencoder (RAE)
\citep{zheng2026diffusion}. 
We use the publicly available \texttt{RAE-dinov2-wReg-base-ViTXL-n08} checkpoint and keep both the encoder and decoder frozen. Input images are bicubically resized and center-cropped
to $256\times256$, producing a $768\times16\times16$ latent representation.

\begin{table}[ht]
\centering
\caption{Experimental settings for the pretrained RAE extension.}
\label{tab:rae_setting}
\small
\begin{tabular}{ll}
\hline
Setting & Value \\
\hline
RAE checkpoint &
DINOv2-based RAE, ViT-XL decoder \\
Input resolution &
$256\times256$ \\
Latent shape &
$768\times16\times16$ \\
RAE parameters &
Frozen \\
Dataset &
ImageNet \\
Predictor train / val / test &
20,000 / 2,000 / 5,000 \\
Predictor hidden channels &
128 \\
Residual blocks &
5 \\
Optimizer &
AdamW \\
Learning rate &
$5\times10^{-4}$ \\
Weight decay &
$10^{-3}$ \\
Batch size &
64 \\
Max.\ epochs / patience &
30 / 8 \\
Gradient clipping &
1.0 \\
$\gamma$ initialization &
0.5 \\
M4 coefficient $\alpha$ &
1.0 \\
M5 coefficient $\beta$ &
1.0 \\
Evaluated $N_{\rm tr}$ &
1, 2 \\
Metrics &
MSE, PSNR, SSIM, LPIPS, rFID-5k \\
\hline
\end{tabular}
\end{table}

The main experimental settings are summarized in
Table~\ref{tab:rae_setting}.

We use fixed ImageNet100 subsets of 20,000 training, 2,000 validation, and 5,000 test images.
The nonlinear corrector follows the architecture in Appendix~\ref{app:implementation}, with 128 hidden channels and five residual blocks (as in the main ImageNet experiments). 

We compare M1, M2 ($y_2,y_3,y_5$), M4, M5, M6-LO, M6
($N_{\rm tr}=1,2$), and M6-xFree ($N_{\rm tr}=1,2$).
The M4 coefficient is fixed to $\alpha=1$ as in the main experiments, whereas validation selected $\beta=1$ for M5 before final test evaluation.

We report MSE, PSNR, SSIM, LPIPS, and rFID-5k. 
The first four metrics are computed per sample.
LPIPS is computed using the pretrained AlexNet-based LPIPS model
\citep{zhang2018unreasonable}, with images normalized to $[-1,1]$, and averaged over the 5,000 test images.
rFID-5k is computed between the same fixed 5,000 ground-truth and reconstructed images using CleanFID in \texttt{legacy\_tensorflow} mode. 
These rFID values are intended for comparisons within this experiment and should not be directly compared with rFID values computed using a different number of images.
For M6 versus M6-LO, 95\% confidence intervals for mean metric differences are computed using paired $t$ intervals over the same 5,000 test images.

% D.2
\subsection{Distinct Latent Geometry of SRD in the pretrained RAE}
\label{app:RAE_SRD}
To examine whether the latent SRD observed in the convolutional AEs persists in a substantially different representation space, we analyze repeated self-reconstruction trajectories of the pretrained RAE. 
Using validation images, we measure the norm of each reverse-step vector $dz_k$, the cosine similarity between successive steps, and the cosine similarity between each step and the initial direction $dz_0$.
As in Sec.~\ref{sec:exp_geometry}, $dz_0$ is used only for geometric analysis and is not available to the $x$-free refinement models.

\begin{figure}[t]
    \centering
    \includegraphics[width=0.5\columnwidth]{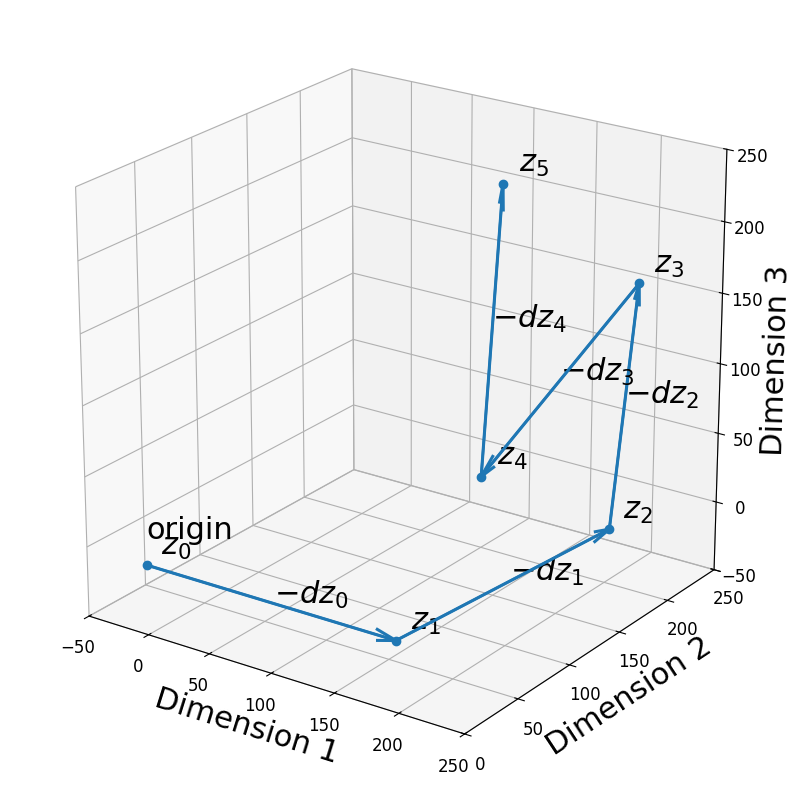}
    \caption{3D visualization of latent SRD trajectories in the pretrained RAE on ImageNet100.}
    \label{fig:RAE_SRD_trajectory}
\end{figure}

\begin{table}[t]
\centering
\caption{
Latent SRD geometry of the pretrained RAE.
Values are mean $\pm$ standard deviation over validation images.
}
\label{tab:rae_geometry}
\small
\begin{tabular}{c|ccccc}
\hline
$k$ & 0 & 1 & 2 & 3 & 4 \\
\hline
$\|dz_k\|_2$
& $197.83{\pm}27.02$
& $184.87{\pm}29.76$
& $192.06{\pm}34.16$
& $200.25{\pm}36.00$
& $203.22{\pm}33.75$ \\
$\cos(dz_k,dz_{k+1})$
& $0.094{\pm}0.067$
& $0.173{\pm}0.073$
& $0.173{\pm}0.081$
& $0.163{\pm}0.088$
& -- \\
$\cos(dz_0,dz_k)$
& --
& $0.094{\pm}0.067$
& $-0.084{\pm}0.043$
& $-0.089{\pm}0.034$
& $-0.072{\pm}0.028$ \\
\hline
\end{tabular}
\end{table}
Table~\ref{tab:rae_geometry} summarizes the latent SRD statistics.
To visualize the corresponding trajectory geometry, we fit a three-dimensional trajectory to the measured step magnitudes and cosine similarities, following the same procedure as in the main experiments, as illustrated in Fig.~\ref{fig:RAE_SRD_trajectory}.
Although the two-dimensional fitting resulted in relatively large errors, the three-dimensional fit reproduces all measured cosine similarities with an average error below $10^{-10}$.
Thus, the figure provides an accurate three-dimensional representation of the measured trajectory geometry.

%Table A4

The RAE exhibits a markedly different SRD geometry from the convolutional AEs in the main experiments.

First, the step magnitude does not progressively decay: after decreasing
from $\|dz_0\|_2=197.8$ to $\|dz_1\|_2=184.9$, it increases again and
reaches 203.2 at $dz_4$, showing a markedly different magnitude profile from the convolutional AEs.

Second, successive directions exhibit only weak positive alignment. 
Their mean cosine similarity is $0.094$ for $(dz_0,dz_1)$ and approximately $0.16$--$0.17$ thereafter, in contrast to the strong local directional coherence observed for the convolutional AEs in
Fig.~\ref{fig:SRD_trajectory_a}.
%Fig.~\ref{fig:trajectory_all}.

Third, the trajectory rapidly departs from its initial direction. While $dz_1$ remains weakly aligned with $dz_0$, the mean cosine similarity with $dz_0$ becomes negative from $dz_2$ onward, reaching $-0.084$, $-0.089$, and $-0.072$ for $dz_2$, $dz_3$, and $dz_4$, respectively.

Taken together, SRD persists in the pretrained RAE but has a qualitatively different geometry: instead of the locally coherent and gradually curving trajectories observed in the convolutional AEs, the RAE exhibits weak local directional alignment, rapid departure from the initial direction, and non-decaying step magnitudes. 
This strongly curved geometry also provides a natural explanation for why fixed linear latent extrapolation is less effective in the RAE, as examined next.

% D.3
\subsection{SRD-Guided Reconstruction Refinement in the Pretrained RAE}
\label{app:RAE_refinement}

% Table A5
% Table A5
\begin{table}[t]
\centering
\caption{
Reconstruction performance of SRD-based refinement on the pretrained RAE.
MSE, PSNR, SSIM, and LPIPS are averaged over the fixed 5,000 test images;
rFID-5k is computed over the same complete test set.
Best values are shown in bold.
}
\label{tab:rae_full_results}
\small
\setlength{\tabcolsep}{4.5pt}
\begin{tabular}{lcccccr}
\hline
Model & $N_{\rm tr}$ & MSE $\downarrow$ & PSNR $\uparrow$
& SSIM $\uparrow$ & LPIPS $\downarrow$ & rFID-5k $\downarrow$ \\
\hline
M1                  & -- & 0.01602 & 18.87 & 0.4696 & \textbf{0.1558} & \textbf{3.91} \\
\hline
M2 ($y_2$)          & 1  & 0.02124 & 17.41 & 0.4286 & 0.2193 & 6.29 \\
M2 ($y_3$)          & 2  & 0.02612 & 16.37 & 0.4002 & 0.2770 & 11.21 \\
M2 ($y_5$)          & 4  & 0.03516 & 14.97 & 0.3601 & 0.3740 & 33.44 \\
\hline
M4                  & 1  & 0.02153 & 17.70 & 0.4245 & 0.2009 & 5.72 \\
M5                  & 1  & 0.01911 & 17.94 & 0.4363 & 0.1943 & 4.74 \\
\hline
M6-LO               & 0  & 0.01584 & 18.65 & 0.4899 & 0.4701 & 25.52 \\
M6                  & 1  & \textbf{0.01483} & \textbf{18.99}
                    & \textbf{0.4948} & 0.4446 & 21.76 \\
M6                  & 2  & 0.01494 & 18.94 & 0.4927 & 0.4527 & 22.66 \\
M6-xFree                & 1  & 0.01791 & 18.20 & 0.4532 & 0.2083 & 5.72 \\
M6-xFree                & 2  & 0.01800 & 18.17 & 0.4524 & 0.2108 & 6.00 \\
\hline
\end{tabular}
\end{table}
Table~\ref{tab:rae_full_results} summarizes the reconstruction results
on the pretrained RAE.
Repeated self-reconstruction again progressively degrades fidelity: from M1 to $y_5$, PSNR decreases from 18.87 to 14.97\,dB, while rFID-5k increases from 3.91 to 33.44.
Thus, the degradation under repeated application observed for the convolutional AEs also occurs in the pretrained RAE, despite its substantially different encoder--decoder architecture and latent representation.
Because PSNR is computed independently for each sample before averaging, its ordering need not exactly match that of the dataset-averaged MSE.

The linear SRD baselines M4 and M5 do not improve over M1, achieving 17.70 and 17.94\,dB, respectively, compared with 18.87\,dB for M1.
The limited effectiveness of fixed linear latent extrapolation is consistent with the weak local alignment and rapid directional changes of the RAE trajectory observed in Appendix~\ref{app:RAE_SRD}.

In contrast, nonlinear SRD conditioning remains informative.
The trajectory-free M6-LO achieves 18.65\,dB, whereas adding the first observable SRD transition improves PSNR to 18.99\,dB for M6 ($N_{\rm tr}=1$) and reduces mean MSE from 0.01584 to 0.01483.
The mean paired differences (M6 - M6-LO) are
$\Delta\mathrm{PSNR}=+0.341\,\mathrm{dB}$
(95\% CI: $[0.310,0.373]$) and
$\Delta\mathrm{LPIPS}=-0.0255$
(95\% CI: $[-0.0262,-0.0248]$).
M6 also improves SSIM and rFID-5k relative to M6-LO
(Table~\ref{tab:rae_full_results}).
Thus, despite the qualitatively different latent dynamics, explicitly providing SRD information consistently improves prediction over the matched trajectory-free baseline.
A second transition provides no further gain, suggesting that the useful SRD horizon depends on the underlying encoder--decoder representation.

M6-xFree shows a complementary result.
With $N_{\rm tr}=1$, it improves PSNR from 17.70\,dB for its SRD-derived linear pseudo-target baseline M4 to 18.20\,dB, showing that SRD-conditioned latent correction can recover part of the loss introduced by the pseudo-target while remaining decoder-compatible.
It remains below M1 in pixel fidelity, and an additional transition again provides no further benefit.
The distinct pixel--perceptual behavior of M6 and M6-xFree is analyzed next.

%D.4
\subsection{Pixel--Perceptual Trade-off and Qualitative Analysis}
\label{app:RAE_perceptual}
The pretrained RAE reveals a clear distinction between the usefulness of SRD information and the perceptual effect of the refinement objective.
As shown above, adding SRD to M6-LO improves all five evaluated metrics.
However, comparison with the original one-shot reconstruction M1 reveals a different issue: optimizing the latent correction for pixel-space MSE does not necessarily preserve perceptual reconstruction quality.

For M6 ($N_{\rm tr}=1$), mean PSNR improves from 18.87 to 18.99\,dB relative to M1, MSE decreases from 0.01602 to 0.01483, and SSIM increases from 0.4696 to 0.4948.
In contrast, both perceptual metrics worsen substantially:
LPIPS increases from 0.1558 to 0.4446, and rFID-5k from 3.91 to 21.76.
Thus, improved pixel fidelity under latent refinement does not necessarily imply improved perceptual similarity in the pretrained RAE.
%
% Importantly, this perceptual degradation is not introduced by SRD conditioning itself.
% The trajectory-free M6-LO exhibits even worse LPIPS and rFID-5k (0.4701 and 25.52) than M6 (0.4446 and 21.76).
% Therefore, adding SRD actually improves both perceptual metrics under the same pixel-MSE refinement objective.
The mismatch also occurs without SRD conditioning: M6-LO has worse LPIPS and rFID-5k than M6 under the same pixel-MSE objective.
Adding SRD therefore improves both perceptual metrics relative to the trajectory-free baseline.
%
%Thus, pixel-MSE-based latent refinement can improve pixel fidelity while degrading the perceptual quality of the original RAE reconstruction.

M6-xFree exhibits a substantially different trade-off.
Although its PSNR is lower than that of supervised M6 (18.20 vs.\ 18.99\,dB), its LPIPS and rFID-5k are much closer to M1 (0.2083 and 5.72, respectively).
% Moreover, relative to its pseudo-target baseline M4, M6-xFree improves PSNR from 17.70 to 18.20\,dB while largely preserving its perceptual characteristics: rFID-5k remains 5.72 and LPIPS changes only from 0.2009 to 0.2083.
It also largely preserves the perceptual characteristics of its
pseudo-target M4: rFID-5k remains 5.72, while LPIPS changes from
0.2009 to 0.2083.
Thus, the SRD-derived pseudo-target leads to a more conservative refinement that preserves perceptual structure substantially better than direct ground-truth pixel-MSE optimization.

\begin{figure*}[t]
    \centering
    % \begin{subfigure}[t]{0.32\textwidth}
    %     \centering
    %     \includegraphics[width=\linewidth]{images/RAE_dPSNR_dLPIPS_M6-LO.png}
    %     \caption{M6-LO}
    %     \label{fig:psnr_lpips_a}
    % \end{subfigure}
    % \hfill
    % \begin{subfigure}[t]{0.32\textwidth}
    %     \centering
    %     \includegraphics[width=\linewidth]{images/RAE_dPSNR_dLPIPS_M6-Ntr1.png}
    %     \caption{M6}
    %     \label{fig:psnr_lpips_b}
    % \end{subfigure}
    % \hfill
    % \begin{subfigure}[t]{0.32\textwidth}
    %     \centering
    %     \includegraphics[width=\linewidth]{images/RAE_dPSNR_dLPIPS_M6-D(Ntr1).png}
    %     \caption{M6-xFree}
    %     \label{fig:psnr_lpips_c}
    % \end{subfigure}
    \includegraphics[width=\columnwidth]{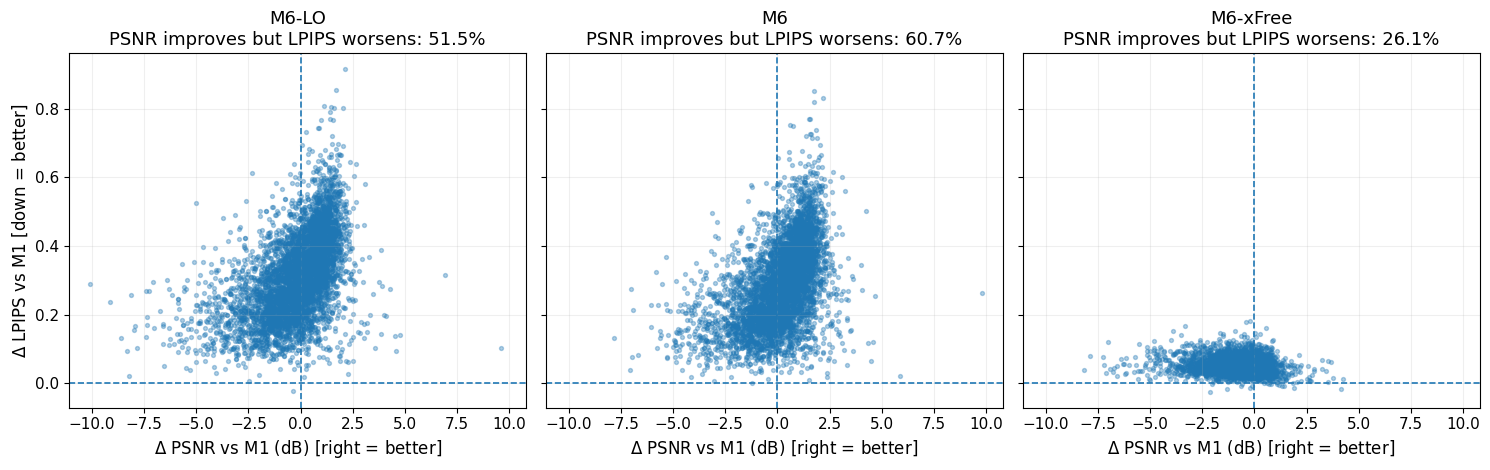}
    \caption{Per-sample pixel--perceptual discrepancy under RAE refinement. \\ Positive $\Delta$PSNR indicates improved pixel fidelity, whereas positive $\Delta$LPIPS indicates degraded perceptual similarity. The frequent occurrence of samples with improved PSNR but worse LPIPS shows that pixel-level improvement does not necessarily imply perceptual improvement. M6 and M6-xFree use $N_{\mathrm{tr}}=1$.
    }
    \label{fig:psnr_lpips}
\end{figure*}
Figure~\ref{fig:psnr_lpips} shows that improved pixel fidelity is
frequently accompanied by worse perceptual similarity at the sample level.
A similar pattern is observed for M6-LO, indicating that this behavior is not specific to SRD conditioning.
%
% The discrepancy therefore reflects a general property of the evaluated pixel-MSE latent refinement rather than a failure specific to SRD conditioning.

\begin{figure*}%[t]
    \centering
    \captionsetup{justification=centering}
    % ===== Top row =====
    \begin{subfigure}[t]{0.32\textwidth}
        \centering
        \includegraphics[width=\linewidth]{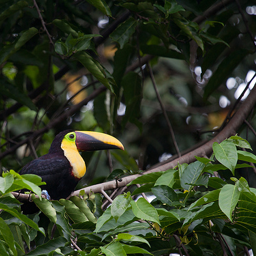}
        \caption{GT}
        \label{fig:RAE_1_gt}
    \end{subfigure}
    \hfill
    \begin{subfigure}[t]{0.32\textwidth}
        \centering
        \includegraphics[width=\linewidth]{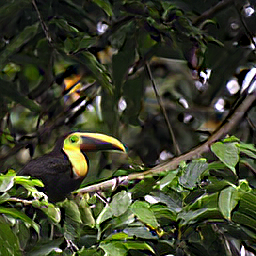}
        \caption{M4 \\ PSNR=14.89dB \\ LPIPS=0.2188}
        \label{fig:RAE_1_m4}
    \end{subfigure}
    \hfill
    \begin{subfigure}[t]{0.32\textwidth}
        \centering
        \includegraphics[width=\linewidth]{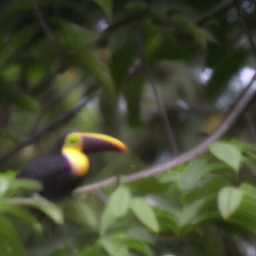}
        \caption{M6-LO \\ PSNR=16.66dB \\ LPIPS=0.4711}
        \label{fig:RAE_1_m6-lo}
    \end{subfigure}

    \vspace{1mm}

    % ===== Bottom row =====
    \begin{subfigure}[t]{0.32\textwidth}
        \centering
        \includegraphics[width=\linewidth]{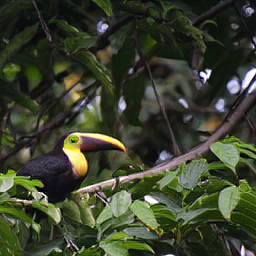}
        \caption{M1 \\ PSNR=16.41dB \\ LPIPS=0.1788}
        \label{fig:RAE_1_m1}
    \end{subfigure}
    \hfill
    \begin{subfigure}[t]{0.32\textwidth}
        \centering
        \includegraphics[width=\linewidth]{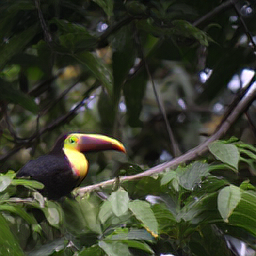}
        \caption{M6-xFree \\ PSNR=16.49dB \\ LPIPS=0.2297}
        \label{fig:RAE_1_m6d}
    \end{subfigure}
    \hfill
    \begin{subfigure}[t]{0.32\textwidth}
        \centering
        \includegraphics[width=\linewidth]{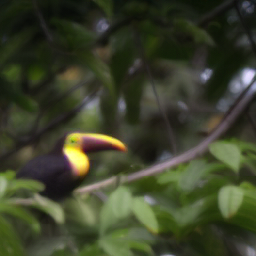}
        \caption{M6\\ PSNR=17.51dB \\ LPIPS=0.4545}
        \label{fig:RAE_1_m6}
    \end{subfigure}
    \caption{
    % Qualitative comparison of reconstruction results from ImageNet.
    % \\(PSNR decreases and LPIPS worsens, while M6-xFree improves PSNR)
    Qualitative comparison on ImageNet100 using the pretrained RAE. \\ M6 improves PSNR but worsens LPIPS relative to M1, while M6-xFree retains more fine-scale detail and achieves lower LPIPS than M6.
    }
    \label{fig:RAEimage_examples1}
\end{figure*}
\begin{figure*}%[t]
    \centering
    \captionsetup{justification=centering}
    % ===== Top row =====
    \begin{subfigure}[t]{0.32\textwidth}
        \centering
        \includegraphics[width=\linewidth]{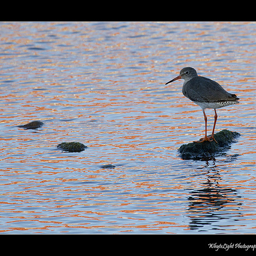}
        \caption{GT}
        \label{fig:RAE_2_gt}
    \end{subfigure}
    \hfill
    \begin{subfigure}[t]{0.32\textwidth}
        \centering
        \includegraphics[width=\linewidth]{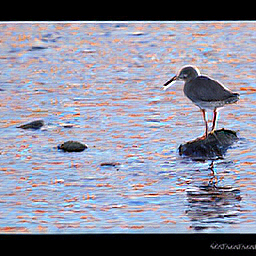}
        \caption{M4 \\ PSNR=15.73dB \\ LPIPS=0.1896}
        \label{fig:RAE_2_m4}
    \end{subfigure}
    \hfill
    \begin{subfigure}[t]{0.32\textwidth}
        \centering
        \includegraphics[width=\linewidth]{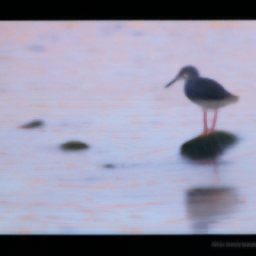}
        \caption{M6-LO \\ PSNR=17.38dB \\ LPIPS=0.4901}
        \label{fig:RAE_2_m6-lo}
    \end{subfigure}

    \vspace{1mm}

    % ===== Bottom row =====
    \begin{subfigure}[t]{0.32\textwidth}
        \centering
        \includegraphics[width=\linewidth]{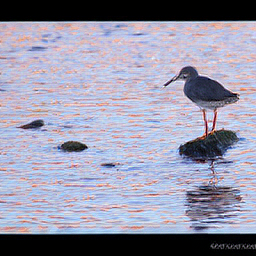}
        \caption{M1 \\ PSNR=15.92dB \\ LPIPS=0.1480}
        \label{fig:RAE_2_m1}
    \end{subfigure}
    \hfill
    \begin{subfigure}[t]{0.32\textwidth}
        \centering
        \includegraphics[width=\linewidth]{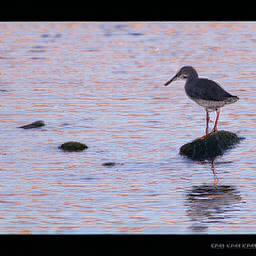}
        \caption{M6-xFree \\ PSNR=17.35dB \\ LPIPS=0.1688}
        \label{fig:RAE_2_m6d}
    \end{subfigure}
    \hfill
    \begin{subfigure}[t]{0.32\textwidth}
        \centering
        \includegraphics[width=\linewidth]{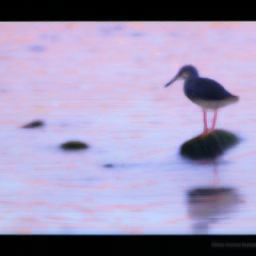}
        \caption{M6\\ PSNR=15.28dB \\ LPIPS=0.4767}
        \label{fig:RAE_2_m6}
    \end{subfigure}
    \caption{
    % Qualitative comparison of reconstruction results from ImageNet.
    % \\(PSNR recrease and LPIPS increase, but M6-xFree showed better PSNR)
    Qualitative comparison on ImageNet100 using the pretrained RAE. \\
M6 deteriorates in both PSNR and LPIPS relative to M1.\\
In contrast, M6-xFree improves PSNR over M1 with a modest increase
in LPIPS and substantially better perceptual similarity than M6.
    }
    \label{fig:RAEimage_examples2}
\end{figure*}
The qualitative examples in Figs.~\ref{fig:RAEimage_examples1} and \ref{fig:RAEimage_examples2} illustrate the same behavior.
M6 can reduce pixel error while smoothing fine textures and structural detail, whereas M6-xFree generally retains more of the perceptual structure present in M1.
These results separate two conclusions: SRD remains useful for predicting latent corrections in the pretrained RAE, but the objective used to exploit this information determines whether improved pixel fidelity translates into perceptually desirable reconstruction.

\end{document}